%% file: main.tex
\documentclass[journal]{IEEEtran}
\usepackage{graphicx}
\usepackage{amssymb}
\usepackage{multirow}
\usepackage{amsmath}
\usepackage{tablefootnote}
\usepackage{color}
\usepackage{xcolor}
\usepackage{comment}
\graphicspath{ {img/} }

\DeclareMathOperator*{\argmin}{arg\,min}
\newcommand{\etal}{\textit{et al. }} % I added an empty space after "al."
\newcommand{\comm}[1]{}

\renewcommand*{\thesubsection}{\Alph{subsection}}

\begin{document}
%IEEE Transactions on Intelligent Transportation Systems
%IEEE Transactions on Vehicular Technology

\title{Cooperative Saliency-based Obstacle Detection and AR Rendering for Increased Situational Awareness}
%\title{Enhancing Driver's Awareness by Obstacle Detection and Scene Augmentation in a Cooperative Setting}
%\title{Road Scene Augmentation by Cooperative Localization of Obstacles} %  for situational awareness

\author{Gerasimos~Arvanitis,
        Nikolaos~Stagakis,
        Evangelia I.~Zacharaki,
        and~Konstantinos~Moustakas,~\IEEEmembership{Senior~Member,~IEEE}% <-this % stops a space
\thanks{G. Arvanitis, N. Stagakis, E. I. Zacharaki and K. Moustakas are with the Department of Electrical and Computer Engineering, University of Patras, Greece (e-mail: arvanitis@ece.upatras.gr, nick.stag@ece.upatras.gr, ezachar@upatras.gr, moustakas@ece.upatras.gr)} % stops a space
%\thanks{This work has received funding from the European Union’s Horizon 2020 research and innovation programme under Grant Agreement No 871738 - CPSoSaware: Crosslayer cognitive optimization tools $\&$ methods for the lifecycle support of dependable CPSoS.}
% <-this % stops a space
%\thanks{Manuscript received December 10, 2021; revised Month XX, 2022.}
}

% The paper headers
\markboth{Journal of \LaTeX\ Class Files,~Vol.~14, No.~8, August~2022}%
{Shell \MakeLowercase{\textit{et al.}}: Bare Demo of IEEEtran.cls for IEEE Journals}

% make the title area
\maketitle
% As a general rule, do not put math, special symbols or citations
% in the abstract or keywords.
\begin{abstract}
Autonomous vehicles are expected to operate safely in real-life road conditions in the next years. Nevertheless, unanticipated events such as the existence of unexpected objects in the range of the road, can put safety at risk. The advancement of sensing and communication technologies and Internet of Things may facilitate the recognition of hazardous situations and information exchange in a cooperative driving scheme, providing new opportunities for the increase of collaborative situational awareness. Safe and unobtrusive visualization of the obtained information may nowadays be enabled through the adoption of novel Augmented Reality (AR) interfaces in the form of windshields. Motivated by these technological opportunities, we propose in this work a saliency-based distributed, cooperative obstacle detection and rendering scheme for increasing the driver's situational awareness through (i) automated obstacle detection, (ii) AR visualization and (iii) information sharing (upcoming potential dangers) with other connected vehicles or road infrastructure. An extensive evaluation study using a variety of real datasets for pothole detection showed that the proposed method provides favorable results and features compared to other recent and relevant approaches.
\end{abstract}

% Note that keywords are not normally used for peer review papers.

% For peer review papers, you can put extra information on the cover
% page as needed:
% \ifCLASSOPTIONpeerreview
% \begin{center} \bfseries EDICS Category: 3-BBND \end{center}
% \fi
%
% For peerreview papers, this IEEEtran command inserts a page break and
% creates the second title. It will be ignored for other modes.
\IEEEpeerreviewmaketitle

\section{Introduction}
\IEEEPARstart{I}{n}formation-centric technologies have started to play a central role in the recent automotive industry boosting new research trends in semi or fully \textit{Automated Driving Systems} (ADS). Autonomous vehicles, ranging from level 3 to level 5 of autonomy \cite{taxonomy}, are expected to operate safely in real-life road conditions, but the reality is that obstacles like potholes, bumps, and other unexpected objects are not uncommon in an everyday driving context. For this reason, the detection and identification of obstacles are imperative for reliable operation of autonomous vehicles \cite{8691693}. 

Moreover, driver inattentiveness plays a major role in driving safety and is the culprit of road accidents around the world \cite{distracted1, distracted2}, thus a lot of work has been devoted in the quantification of the abstract mechanics of human situational awareness \cite{autorate}. Enhancing situational awareness is especially critical in the case of semi-autonomous cars, where the operator may be distracted by secondary activities, e.g. looking at the phone or reading a book. If the driver has to take over control, it is important to minimize the required reaction time. This can be achieved by monitoring and presenting to the driver the crucial information about the environment, thus keeping him/her aware of potentially hazardous situations. Inherent challenges include the need for unobtrusive information display, avoiding the effects of tunnel vision which could lead to actually overlooking critical information \cite{9766081}. 

The problem of road pothole detection is commonly targeted using imaging (camera) data and computer vision techniques \cite{8788687, 9676673, 8809907}. Although image-based techniques have achieved great success, one common drawback is that they are sensitive to motion blur and changes in lighting and/or even shadows \cite{9432832}. Also, most techniques do not account for other passing vehicles \cite{9043541}. This can make them unreliable in real use cases, which is a major weakness in problems involving human safety. In light of all this, the use of a 3D LiDAR (Light Detection and Ranging) sensor could provide more robust sensing capabilities for the analysis of potholes, in the same way that it is used to increase the accuracy of road's boundary detection \cite{8654619, 9205694, 8642507}. On the other hand, a limitation of the LiDAR sensor is that, due to refraction and reflection, water appears as a black hole in the \text{imagery} calculated from LiDAR data \cite{chen2017}, imposing additional challenges in the detection of potholes filled with water.

The purpose of this work is to increase the driver's situational awareness through automated cooperative obstacle detection, visualization and information sharing with other connected vehicles in a V2X (vehicle-to-everything) setting. To address the above issues, we developed a point cloud processing system that takes as input road environment data and classifies them into safe and potentially hazardous regions by identifying obstacles lying in the range of the road. We selected LiDAR as sensing modality for the surrounding environment due to its ability to retrieve depth information and its large range, making it suitable for driving environments. For more robust estimation, LiDAR data are fused with information on driving patterns, such as the steering angle of the wheels. For implementation and evaluation, we utilized the open-source CARLA simulator \cite{carla17} including also a multi-agent system of vehicles, and we augmented it with our obstacle detection and tracking component. In this simulated environment, information sharing between agents is enabled, so that vehicles are notified about incoming obstacles even when there is no direct line-of-sight. 

To avoid any information visualization clutter, we propose the use of AR for visualizing critical information in the driver's field of view. AR rendering is based on classical perspective projection, where for each point (of the point cloud) the pixel coordinates in the image space of the AR interface are calculated through projection and a color is assigned indicating the object class. Interfaces that can be used for in-vehicle visualization include AR headset, Head-Up Display (HUD) \cite{8326307, 9304610} or even the car's windshield with transparent display. % 

The contributions of the proposed approach can be summarized as follows. 
\begin{itemize}
    \item Development of an obstacle detection module that takes into account the extraction of saliency maps from point clouds. 
    \item Generation of data for randomized multi-ego connected vehicle in cooperative driving scenarios.
    \item Creation of realistic synthetic data of potholes that can be entered in the town maps of the CARLA simulator for the design of lifelike  driving situations.
    \item AR visualization for point cloud projection registered on the scene images.
     \item Development of public and open access libraries with code for the aforementioned components\footnote{https://github.com/Stagakis/saliency-from-pointcloud}$^{,}$\footnote{https://github.com/Stagakis/carla-data-generation}$^{,}$\footnote{https://github.com/Stagakis/roadpatch-with-pothole-generator}$^{,}$\footnote{https://github.com/Stagakis/carlapclprocessing}.
\end{itemize}

The rest of this paper is organized as follows. First we present previous works in related domains in Section \ref{sec:previous_work}, and then describe in detail the proposed methodology in Sections \ref{sec:obstacle_detection} and \ref{sec:visualization_communication}. Section \ref{sec:experiments} follows with some experimental results in comparison with other state-of-the-art methods, while Section \ref{sec:conclusions} draws the conclusions and directions for future work.

\input{previous_work}

\section{Obstacle Detection}
\label{sec:obstacle_detection}

%The per frame outputs of the Carla simulator that we use are:
% \begin{enumerate}
%     \item the corresponding point clouds received from the LiDAR device
%     \item the steering angle of the wheels per frame.
%  \end{enumerate}

\begin{figure*}[ht]
\centering
\includegraphics[width=0.9\textwidth]{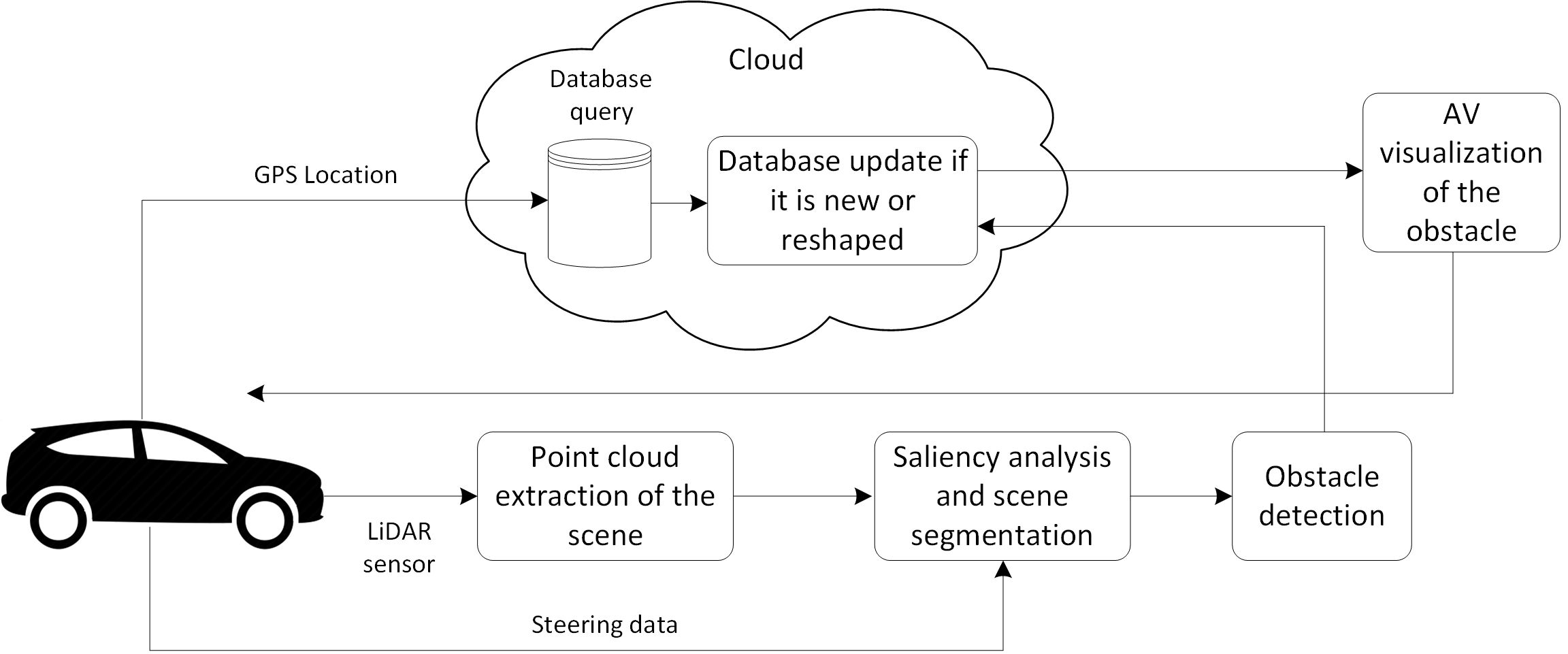} %pipeline.png
\caption{Schematic diagram of the proposed methodology.}
\label{fig:pipeline}
\end{figure*}

This section presents the proposed methodology on obstacle detection and is followed by section \ref{sec:visualization_communication} on visualization and communication aspects.
The main components of the methodology are illustrated in the schematic diagram in Fig.~\ref{fig:pipeline} and can be encapsulated in the next steps:
\begin{itemize}
    \item \textit{Extraction of saliency map}: A saliency value is estimated for any point of the point cloud scene based on its local geometry, as well as the local geometry of its neighboring points.
    \item \textit{Scene segmentation}: The estimated saliency map is then used as a feature to segment the point cloud into areas characterizing (i) the safe area of the road, (ii) be-aware or dangerous areas within the range of the road, and (iii) areas out of the range of the road.
    \item \textit{Static object recognition}: Static objects (i.e., potholes and bumps) can be identified and their point coordinates are stored and then used for the AR-based visualization and communication to other nearby vehicles.
    %\item \textit{AR visualization}: The detected objects are projected in the AR coordinate system as 2D transparencies.
    %\item \textit{Calculation of occupancy factor}: In order to adapt the AR interface to the situational awareness risk, we assume that the required level of attention is propositional to the occupancy of the road in front of (and around) the vehicle, and we, thus, formulate a metric (called \textit{occupancy factor}) that characterizes the amount of obstacles (e.g., cars, potholes, pedestrians) located in the extent of the road, and their relative distance to the vehicle. %%where do we use this step?
\end{itemize}

\begin{figure}[h]
\centering
\includegraphics[width=0.5\textwidth]{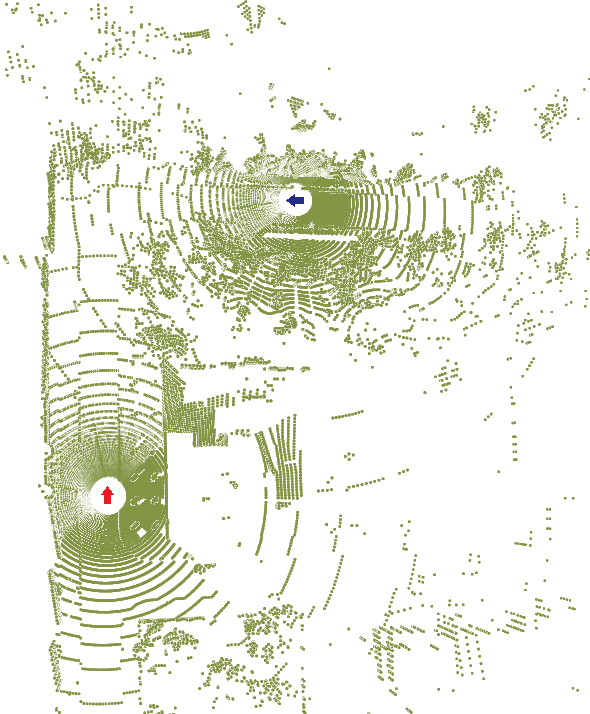}
\caption{Point cloud map of both two vehicles (ego1 and ego2).}
\label{ego0_and_eg01}
\end{figure}

In this work, we assume the existence of two or more vehicles (referred as \textit{ego1} and \textit{ego2} vehicles in this paper) that are moving on the same map of a town but not necessarily at the same time, i.e., they are in spatial proximity but possibly not in temporal proximity. Fig.~\ref{ego0_and_eg01} presents an example of two registered point clouds, as received by the LiDAR devices of ego1 and ego2 vehicles, showing also their starting points (in arrows). We would like to mention here that all the following analysis is applied to each vehicle separately.   

\subsection{Notations}
\label{subsubsec:notations}
Before presenting details on the individual steps, we provide here the necessary definitions and notations. The input data constitute a sequence of point clouds $\mathbf{P}_i$, $i=1,...,l$ that represents a set of $l$ consecutive frames acquired by a LiDAR device. Each point cloud $\mathbf{P}_i$ consists of $m_i$ vertices $\mathbf{v}$, where the value of $m_i$ may be different from frame to frame. The $j$-th vertex ($\mathbf{v}_j$) of a point cloud $\mathbf{P}_i$ is represented by the Cartesian coordinates, denoted $\mathbf{v}_j = \left[x_j,\ y_j,\ z_j\right]^T, \ \forall \ j = 1,\cdots,m_i$, where the index $i$ of the point cloud is omitted for simplification. Thus, all the vertices can be represented as a matrix $\mathbf{V} = \left[\mathbf{v}_1,\ \mathbf{v}_2,\ \cdots,\mathbf{v}_{m_i}\right]\in \mathbb{R}^{3\times m_i}$.
Let's also denote with $\Psi_{j}^{k}$ the set of the indices of the $k$ nearest neighbors of point $j$. For a face $f$ defined by three vertices $(\mathbf{v}_{j1},\mathbf{v}_{j2},\mathbf{v}_{j3})$, the outward unit face normal $\mathbf{n}_{f}$ is calculated by the following equation:
\begin{equation}
\mathbf{n}_{f} = \frac{\left(\mathbf{v}_{j2}-\mathbf{v}_{j1}\right) \times \left(\mathbf{v}_{j3}-\mathbf{v}_{j1}\right)}{\left\|\left(\mathbf{v}_{j2}-\mathbf{v}_{j1}\right) \times \left(\mathbf{v}_{j3}-\mathbf{v}_{j1}\right)\right\|}
\label{face_normals}
\end{equation}
The point normal $\mathbf{n}_{j}$, representing the normal of each point separately, is calculated as:
\begin{equation}
\mathbf{n}_{j} = \frac{\sum_{\forall \mathbf{n}_{f} \in \Psi_j^k} \mathbf{n}_{f}}{|\Psi_j^k|}
\label{point_normals}
\end{equation}

\subsection{Saliency Map Estimation of the Point Cloud Scene}
The purpose of this step is to calculate a metric of saliency for each vertex of a point cloud. Assuming point clouds without context information, saliency characterizes the geometric properties in a local neighborhood of points, i.e., high saliency values represent more perceptually prominent vertices which usually correspond to sharp corners (high-frequency spatial information). On the opposite, the geometrically least important points are those that lie in flat areas. 

For the estimation of the saliency map, we implemented and modified the fusion technique presented in \cite{arvanitis2021}. Instead of using guided normals of centroids, as in the original version \cite{arvanitis2021}, we now utilize normals for the points. This was performed to accelerate computations. Since the number of faces is usually approximately twice the number of vertices, the point normals are almost half the number of the centroid normals. For the sake of completeness, we present here our approach for the estimation of the saliency map of a point cloud scene, utilizing point normals. 

Our fusion technique combines geometric saliency ($s^{(1)}$) with spectral saliency ($s^{(2)}$) features. The unique characteristics of each of these saliency features make the methodology more robust to point clouds acquired under real conditions, thereby being potentially affected by noise and outliers. The method processes each frame independently without examining past temporal information. Thus, as the methodology is applied for each point cloud in the sequence independently, for simplicity we omit the index $i$ (indicating the frame number) from now on in the equations.
%For the sake of simplicity, we will present the process that we follow for one point cloud and the same steps can be used for any other point cloud frame of the sequence.

%\color{red}
%I can show in a fig. how the points and the normals look %per different geometrical features.
%\color{black}

For a point cloud $\mathbf{P}$ with $m$ vertices, a matrix $\mathbf{E} \in\ \mathbb{R}^{3m\times (k+1)}$ is constructed which includes in the first column the $m$ point normals ($\mathbf{n}_{j} = [{n}_{j_x}, {n}_{j_y}, {n}_{j_z}]^T$) of each vertex $j$, $j=1,\cdots,m$, respectively, and in the subsequent $k$ columns the point normals of the $k$ nearest neighbors of vertex $j$ (i.e. $\mathbf{n}_{j\kappa}\in \Psi_j^k$). %, as shown next:
%\begin{equation}
%%\small
%\mathbf{E} = \begin{bmatrix}
%\mathbf{n}_{1} & \mathbf{n}_{{11}} & \mathbf{n}_{{12}} & \dots  & \mathbf{n}_{{1k}} \\
%\mathbf{n}_{2} & \mathbf{n}_{{21}} & \mathbf{n}_{{22}} & \dots  & \mathbf{n}_{{2k}} \\
%\vdots & \vdots & \vdots & \ddots  & \vdots\\
%\mathbf{n}_{m} & \mathbf{n}_{m1} & \mathbf{n}_{m2} & \dots  & \mathbf{n}_{mk} \\
%\end{bmatrix}  %\forall \ i  \  1, \ \cdots, l
%\label{eq:matrixcreation}
%\end{equation}
The salient features extracted by this approach capture global information since the matrix $\mathbf{E}$ is constructed using the point normals of the whole scene.

In order to exploit the geometrical coherence between neighboring normals, we apply Robust Principal Component Analysis (RPCA) to decompose the matrix $\mathbf{E}$ into a low-rank matrix $\mathbf{L}  \in\ \mathbb{R}^{3m\times (k+1)}$ and a sparse matrix $\mathbf{S} \in\ \mathbb{R}^{3m\times (k+1)}$, as described in the appendix \ref{appendix:RPCA}. The matrix $\mathbf{L}$ consists of the low-rank values $\bar{\mathbf{n}}$ of the point normals $\mathbf{n}$, % represented by:
%\begin{equation}
%\mathbf{L} = \begin{bmatrix}
%\bar{\mathbf{n}}_{1} & %\bar{\mathbf{n}}_{{11}} & %\bar{\mathbf{n}}_{{12}} & \dots  & %\bar{\mathbf{n}}_{{1k}} \\
%\bar{\mathbf{n}}_{2} & %\bar{\mathbf{n}}_{{21}} & %\bar{\mathbf{n}}_{{22}} & \dots  & %\bar{\mathbf{n}}_{{2k}} \\
%\vdots & \vdots & \vdots & \ddots  & %\vdots\\
%\bar{\mathbf{n}}_{{m}} & %\bar{\mathbf{n}}_{{m1}} & %\bar{\mathbf{n}}_{{m2}} & \dots  & %\bar{\mathbf{n}}_{{mk}} \\
%\end{bmatrix}  %\ \forall \ i  \  1, \ %\cdots, f
%\label{eq:lowrankmatrix}
%\end{equation}
while the matrix $\mathbf{S}$ consists of the corresponding sparse values represented as $\dot{\mathbf{n}}$. The values of this matrix are zero (or to be more specific nearly zero) if the row (representing a neighboring patch of points) corresponds to point normals with very similar values, i.e., the vertex lies in a flat area, and very large values if the row corresponds to point normals with big dissimilarity (i.e., the vertex lies in a very sharp corner). The fact that most of the local patches $\Psi_j^k$ of a 3D surface are piecewise flat confirms that the matrix $\mathbf{S}$ can be considered a sparse matrix. 
\begin{equation}
\mathbf{S} = \begin{bmatrix}
\dot{\mathbf{n}}_{1} & \dot{\mathbf{n}}_{{11}} & \dot{\mathbf{n}}_{{12}} & \dots  & \dot{\mathbf{n}}_{{1k}} \\
\dot{\mathbf{n}}_{2} & \dot{\mathbf{n}}_{{21}} & \dot{\mathbf{n}}_{{22}} & \dots  & \dot{\mathbf{n}}_{{2k}} \\
\vdots & \vdots & \vdots & \ddots  & \vdots\\
\dot{\mathbf{n}}_{{m}} & \dot{\mathbf{n}}_{{m1}} & \dot{\mathbf{n}}_{{m2}} & \dots  & \dot{\mathbf{n}}_{{mk}} \\
\end{bmatrix}  %\ \forall \ i  \  1, \ \cdots, f
\label{eq:sparsematrix}
\end{equation}
In other words, sparsity of the matrix is assumed because piecewise flat areas are the most dominant geometrical pattern in a 3D surface. 

\subsubsection{Estimation of the geometrical saliency (global approach)}
\label{geometrical_analysis}
As the similarity of normals between neighboring points is a measure of geometrical coherence of the local neighborhood, we estimate the sparsity of the dissimilarity of normals and use it as a feature for geometrical saliency, $s^{(1)}$.%$s_1$.
Low values of the sparse matrix indicate that the normals of the point and its neighbors are similar (low-rank). This means that if all points in a neighborhood have similar geometrical characteristics, the respective patch represents a flat area. On the opposite, high dissimilarity indicates that the surface has an irregular shape. For a point $\mathbf{v}_j$ the geometric saliency feature, $s_{j}^{(1)}$, is estimated by the values of the first column of the sparse matrix $\mathbf{S}$ according to:
\begin{equation}
%\small
s_{j}^{(1)} = ||\dot{\mathbf{n}}_{j}||^2 = \sqrt{\dot{n}_{j_x}^2 + \dot{n}_{j_y}^2 + \dot{n}_{j_z}^2} \ \ \forall \ j = 1,\cdots, m  %\sqrt{\dot{n}_{j1_x}^2 + \dot{n}_{j1_y}^2 + \dot{n}_{j1_z}^2} \ \ \forall \ j = 1,\cdots, m 
\label{geometric_saliency}
\end{equation}
where $\dot{n}_{j_x}$ denotes the scalar value of the $x$ coordinate, of the $[3\cdot(j-1)+1]^{th}$ row, of the $1^{st}$ column of the $\mathbf{S}$ matrix.

\subsubsection{Estimation of the spectral saliency (local approach)}
\label{spectral_analysis}
For the estimation of the spectral-based saliency, $s_{j}^{(2)}$\comm{$s_{2}$}, for a vertex $j$ of the point cloud, we use the submatrix $\mathbf{E}_{j} \ \in \ \mathbb{R}^{3 \times(k+1)}$, that includes the 3 corresponding rows of the matrix $\mathbf{E}$:% in Eq. \eqref{eq:matrixcreation}:
\begin{equation}
%\small
\mathbf{E}_{j} = \begin{bmatrix}
{n}_{j_x} & {n}_{j_{x1}} & {n}_{j_{x2}} & \dots  & {n}_{j_{xk}} \\
{n}_{j_y} & {n}_{j_{y1}} & {n}_{j_{y2}} & \dots  & {n}_{j_{yk}} \\
{n}_{{j_z}} & {n}_{j_{z1}} & {n}_{j_{z2}} & \dots  & {n}_{j_{zk}} \\
\end{bmatrix},
\begin{array}{c}
 %\forall \ i = 1,\cdots, f \\
 \forall \ j = 1,\cdots, m%m_i 
\end{array}
\label{NN1}
\end{equation}
In other words, each submatrix $\mathbf{E}_{j}$, which is a subset of the global matrix $\mathbf{E}_{i}$, 
consists of the point normals of a local neighborhood of the vertex $\mathbf{v}_j$. Then, for each one of these local matrices $\mathbf{E}_{j}$, the covariance matrix $\mathbf{R}_j \in \mathbb{R}^{3\times 3}$ is calculated: 
\begin{equation}
%\small
\mathbf{R}_{j} = \mathbf{E}_{j}\mathbf{E}_{j}^{T}
\label{NN}
\end{equation}
Next, the calculated matrix $\mathbf{R}_j$ is decomposed into a matrix $\mathbf{U}$ consisting of the eigenvectors and a diagonal matrix $\mathbf{\Lambda} = \text{diag}(\lambda_{j1}, \lambda_{j2}, \lambda_{j3})$ consisting of the corresponding eigenvalues, i.e., 
$[\mathbf{U} \ \  \mathbf{\Lambda}]  = \text{eig}(\mathbf{R}_{j})$,
where eig(.) represents the eigendecomposition operation.%function that returns the matrices $\mathbf{U}$ and  $\mathbf{\Lambda}$.

Finally, the spectral saliency of each vertex is calculated by the inverse $l^2$-norm of the corresponding eigenvalues:
\begin{equation}
%\small
s_{j}^{(2)} =  \frac{1}{\sqrt{\lambda_{j1}^2 + \lambda_{j2}^2 + \lambda_{j3}^2}} \ \forall \ j = 1,\cdots, m 
\label{spectral_saliency}
\end{equation}
Eq. \eqref{spectral_saliency} indicates that large values of the term $\sqrt{\lambda_{i1}^2 + \lambda_{i2}^2 + \lambda_{i3}^2}$ correspond to small saliency features implying that the centroid lies in a flat area, while small values of the eigenvalues' norm correspond to large saliency, characterizing the specific centroid as a discriminative point. 

\begin{figure}
		\centering
		\includegraphics[width=0.99\linewidth]{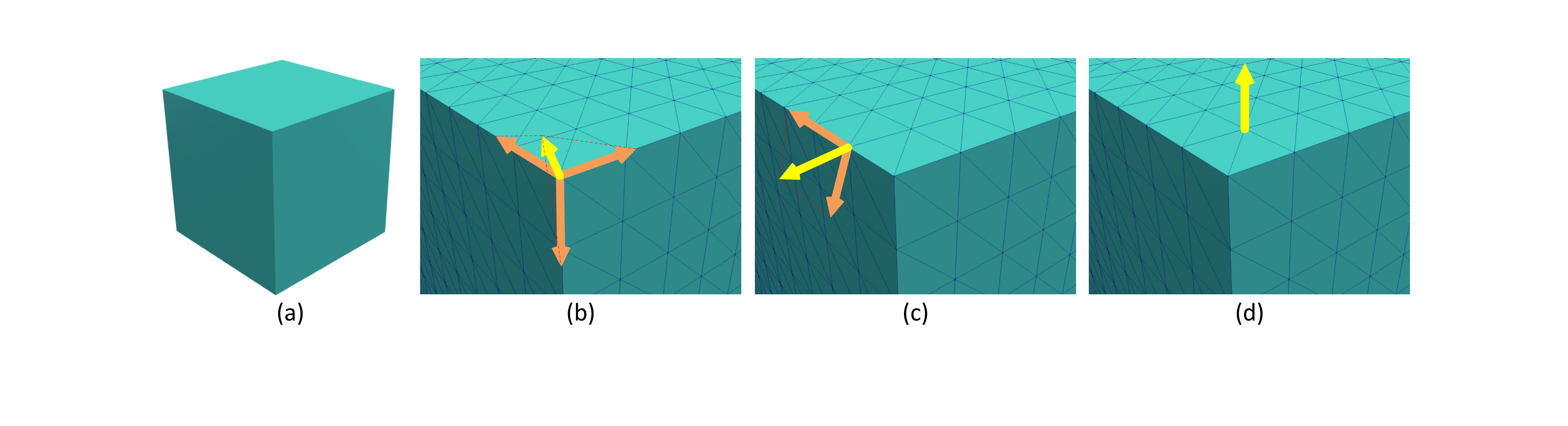}
		\caption{(a) Cube model, (b) corner (\(\lambda_{i1} \cong  \lambda_{i2} \cong  \lambda_{i3}\)), (c) edge (\( \lambda_{i1} \cong  \lambda_{i2} > \lambda_{i3}\)), (d) flat area (\(\lambda_{i1} > \lambda_{i2} \cong  \lambda_{i3}\)).}\label{fig:features_normal}
	\end{figure}

This can be easily justified by the fact that a point normal lying on a flat area is represented by one dominant eigenvector, the corresponding eigenvalue of which has a very large value (especially, considering that it is squared). On the other hand, the point normal of a vertex lying on a corner is represented by three eigenvectors, that correspond to eigenvalues with small and almost equal amplitude, as shown in Fig.~\ref{fig:features_normal}.  

\subsubsection{Normalization and fusion of local and global saliency}
Finally, we linearly scale the values of the geometric ($s^{(1)}$) and spectral ($s^{(2)}$) saliency in the range of [0-1] %, according to:
% \begin{equation}
% \bar{s}_{lj} =  \frac{s_{lj} - \text{min}(s_{lj})}{\text{max}(s_{lj})-\text{min}(s_{lj})} \ \begin{array}{c} 
% \forall \ j = 1,\cdots, m_i , \\ 
% l \ \ \in \{1,2\}
% \end{array}
% \label{norm_spectral_saliency}
% \end{equation}
and combine them through weighted averaging according to:
\begin{equation}
s_{j} = \frac{ w_1\bar{s_{j}}^{(1)} + w_2\bar{s_{j}}^{(2)} }{w_1+w_2} 
\ \ \forall \ j = 1,\cdots, m_i 
\label{combined_saliency}
\end{equation}
where $\bar{s}^{(1)}$ and $\bar{s}^{(2)}$ denote the normalized geometric and spectral saliency features, and $w_1$ and $w_2$ the corresponding weights. We note here that we used equal weights ($w_1 = w_2 = 1$) in all of our experiments, however, the weights can be tuned to emphasize the local or global saliency descriptors, respectively. 

\begin{figure}[h]
\centering
\includegraphics[width=0.5\textwidth]{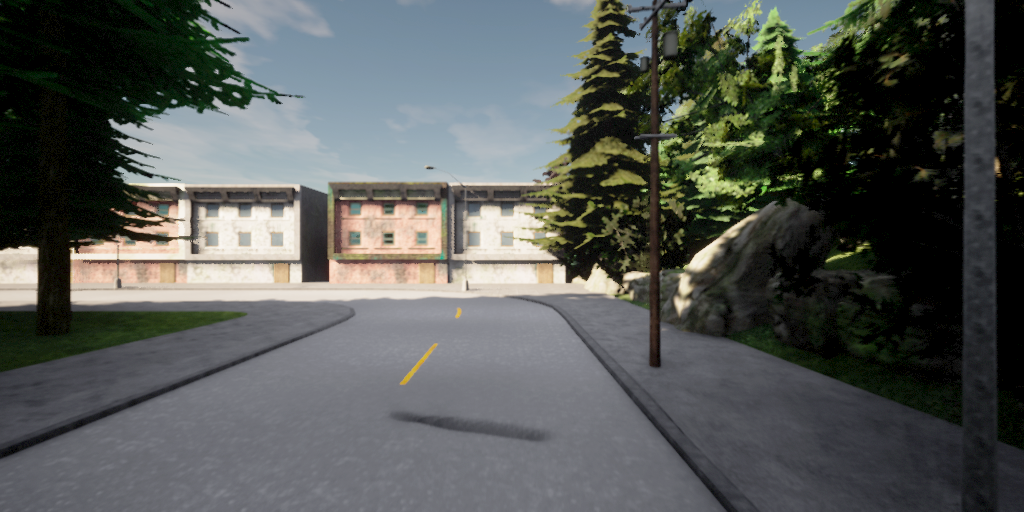}
\caption{Image from the camera of the vehicle, the texture of a pothole is also apparent.}
\label{fig:oneframe}
\end{figure}

\begin{figure}[h]
\centering
\includegraphics[width=0.5\textwidth]{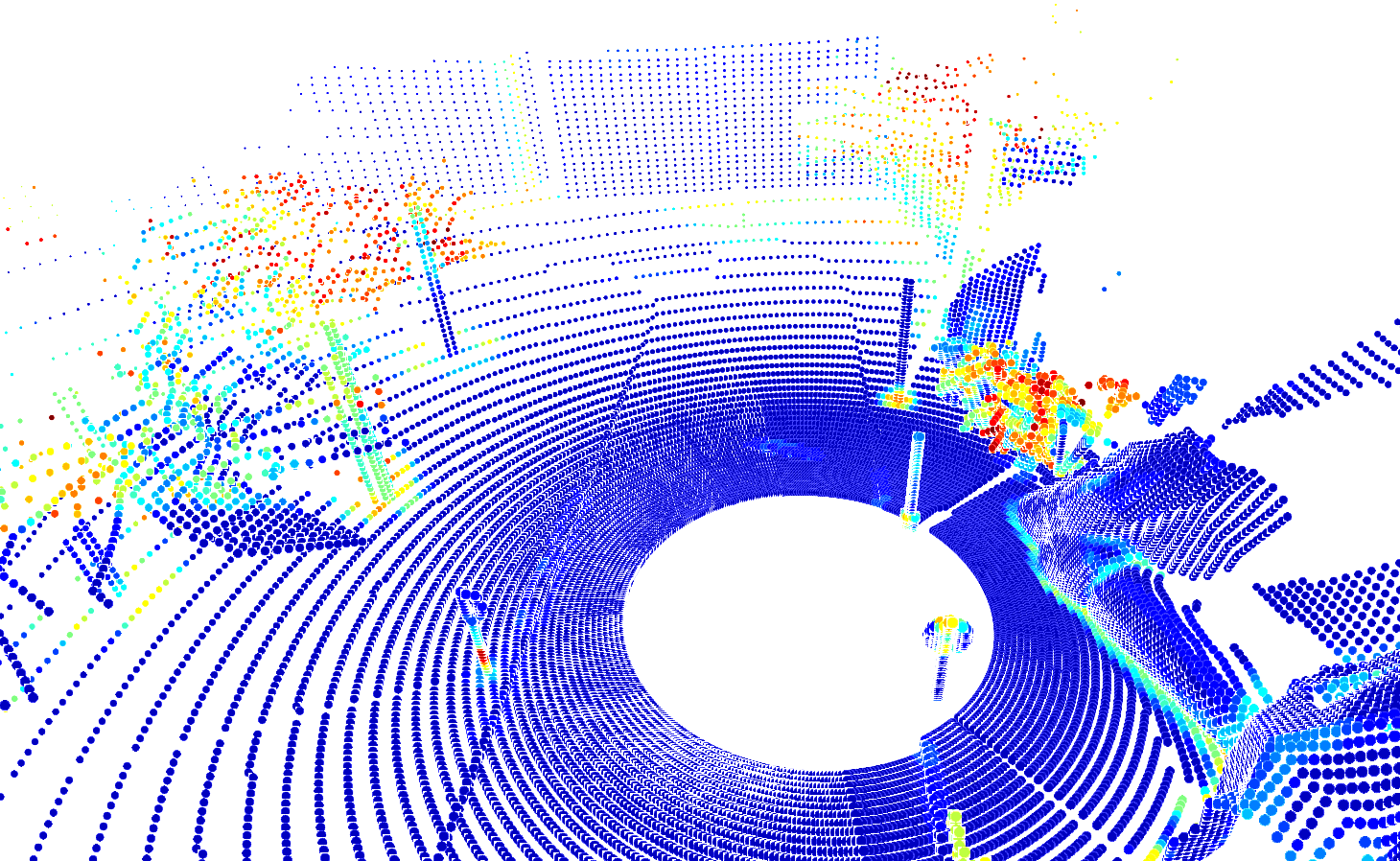}
\caption{Example of saliency map extracted from the road scene shown in Fig.~\ref{fig:oneframe}. }
\label{fig:saliency_mapping}
\end{figure}

The proposed method has shown to be robust \cite{arvanitis2021, arvanitis2019}, even for complex surfaces with different geometrical characteristics and patterns, since it exploits spectral properties (i.e., sensitivity in the variation of neighboring normals) and geometrical characteristics (i.e., sparsity of intense prominent spatial features). An example of the visualization of the saliency map, as applied to the point cloud of a scene shown in Fig.~\ref{fig:oneframe}), is presented in Fig.~\ref{fig:saliency_mapping}.

\subsection{Scene Segmentation for the Identification of On-road Obstacles}

The saliency map of each frame is used to categorize different regions of the scene. For illustration purposes the regions are visualized in different colors:
\begin{itemize}
    \item Blue: The safe area of the road beyond the view of the driver.
    \item Yellow: Be-aware areas representing negative obstacles.
    \item Cyan: Hazardous areas in the range of the road representing positive obstacles.
    \item Purple: Dangerous areas outside of the range of the road.
    \item Red: Recognized obstacles in the range of the road (e.g., potholes).
\end{itemize}

\begin{figure}[h]
\centering
\includegraphics[width=0.5\textwidth]{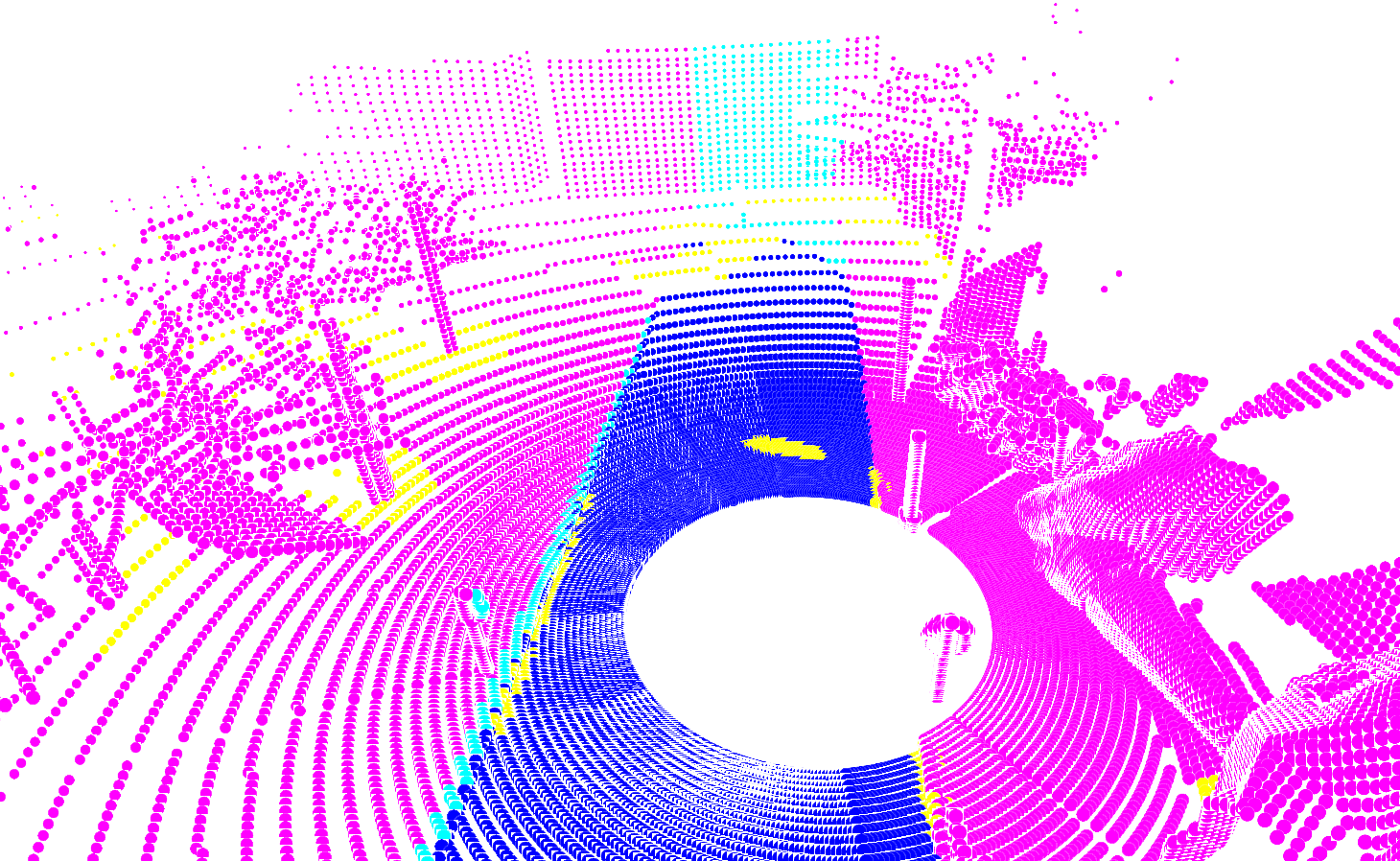}
\caption{Segmentation of the point cloud scene based on the saliency map in Fig.~\ref{fig:saliency_mapping} and the vehicle's moving direction.}
\label{segmentation}
\end{figure}

To define the vehicle's moving direction steering data are used received by internal sensors of the vehicle. The direction of the vehicle specifies which part of the scene in the field of view is in front of the vehicle and is used as as parameter, in addition to saliency mapping, for the segmentation of the point cloud. The more critical regions are the ones that lie within the limits of the road. A segmentation example is illustrated in Fig.~\ref{segmentation}.

%%%%%%%%% 
%% Add an other section before the experimental results explaining the concept and describing the images.
% I have to describe the rest images

\begin{figure}[h]
\centering
\includegraphics[width=0.5\textwidth]{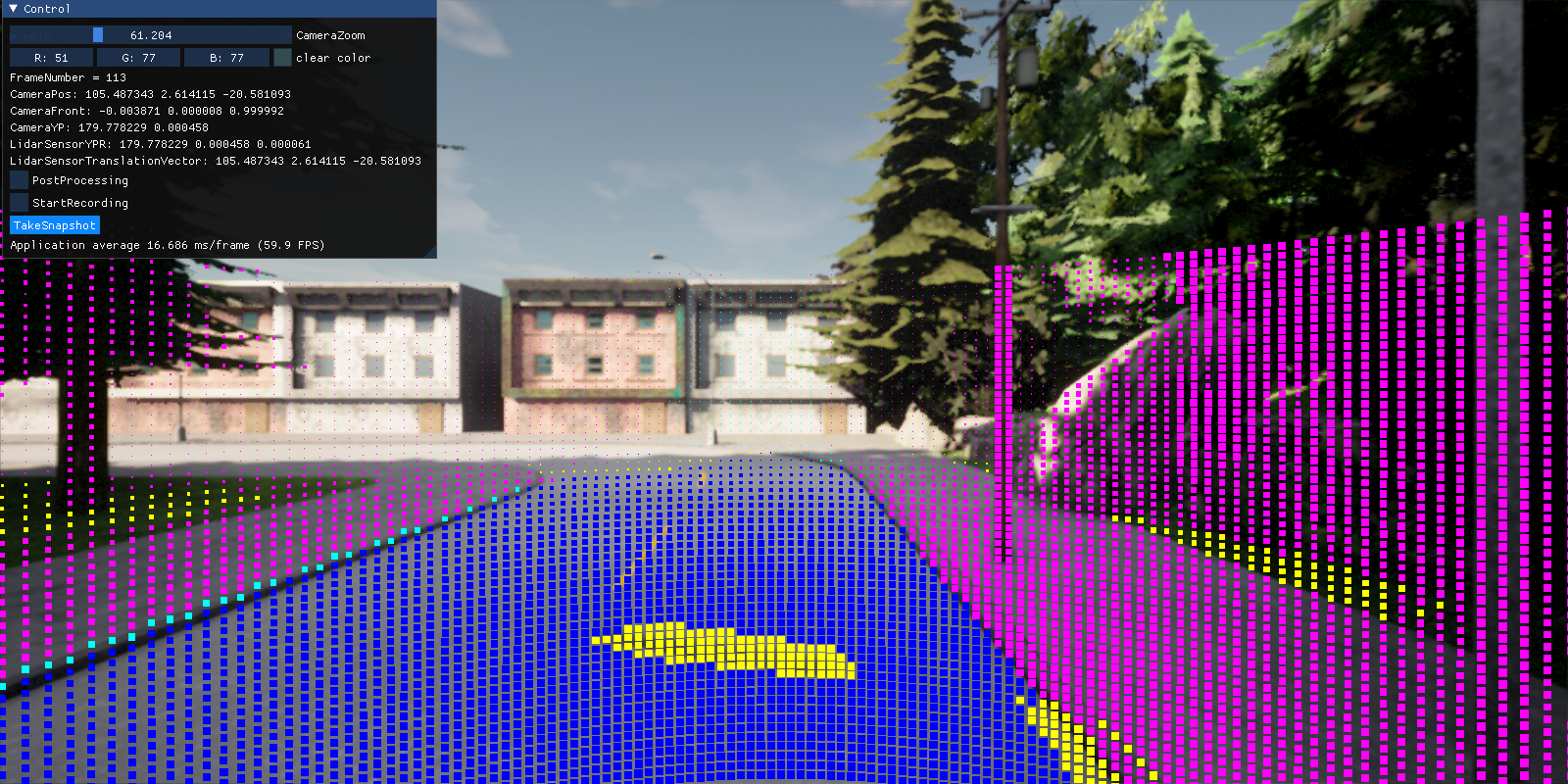}
\caption{Example of segmentation of the point cloud projected to the AR interface (in the view of ego1).}
\label{00113}
\end{figure}

\begin{figure}[h]
\centering
\includegraphics[width=0.5\textwidth]{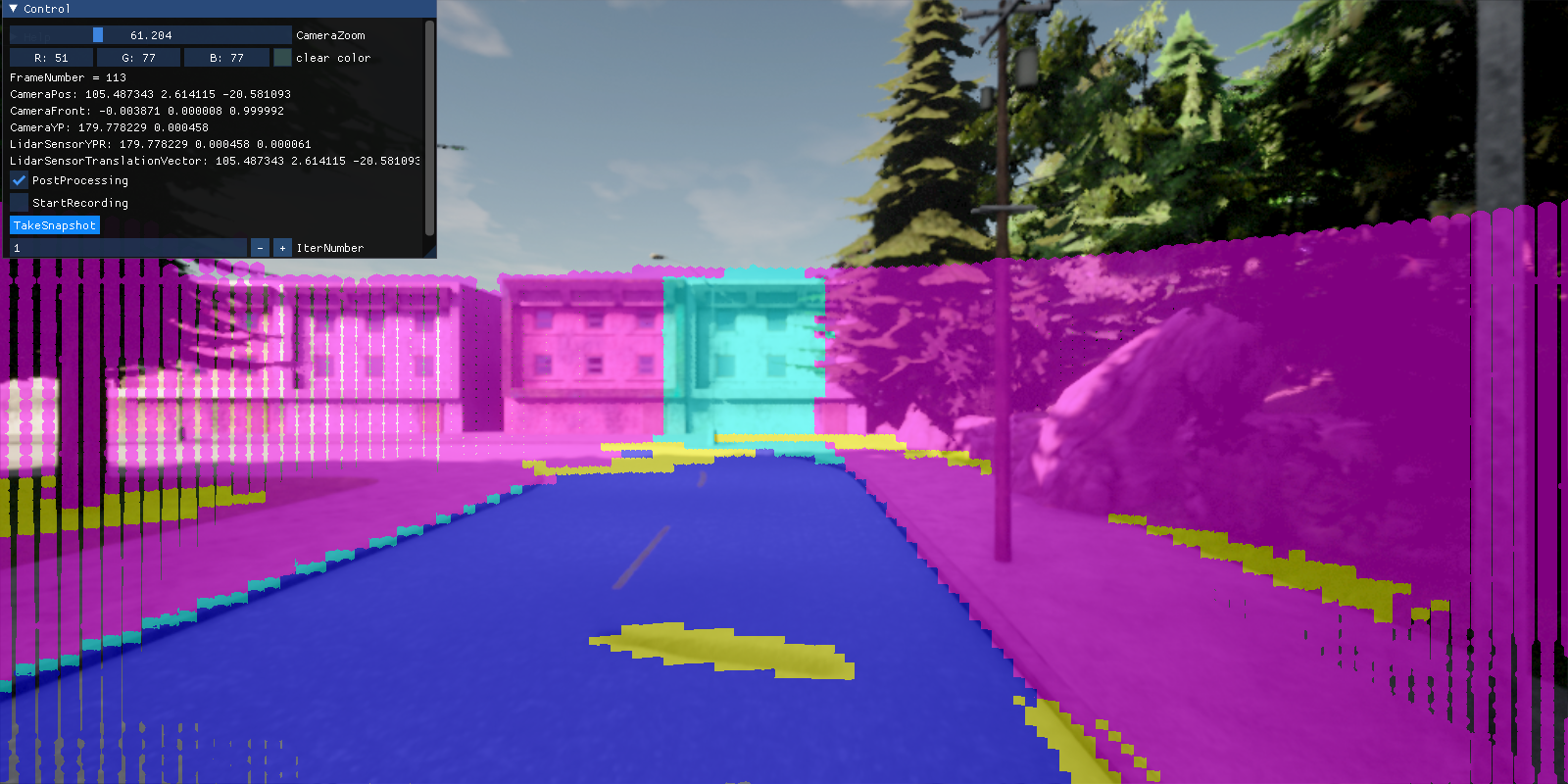}
\caption{Perspective  projection  of  the point cloud vertices to  the  AR  interface  and  image filling (in the view of ego1).}
\label{00113_1}
\end{figure}

\begin{figure}[h]
\centering
\includegraphics[width=0.5\textwidth]{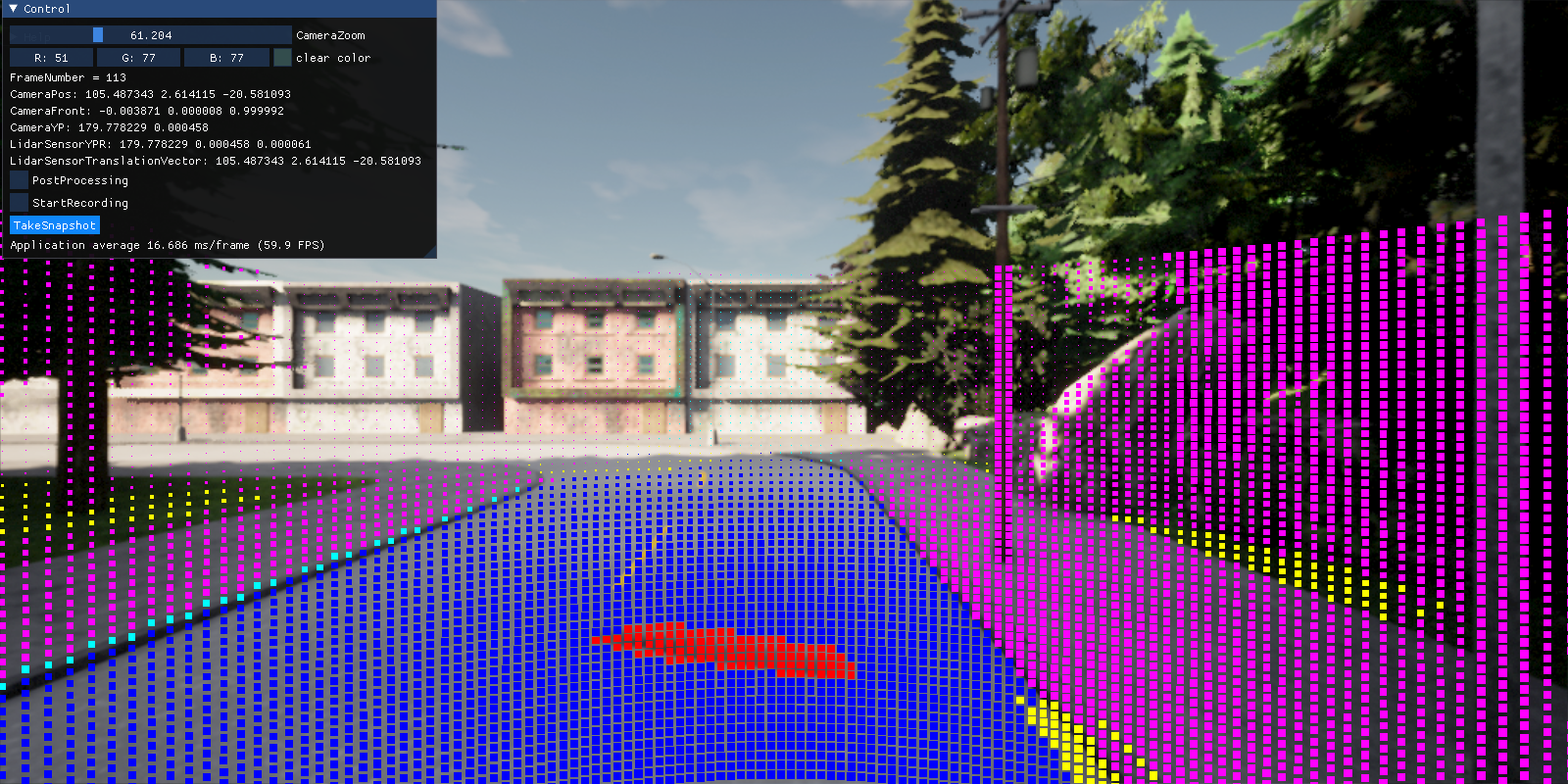}
\caption{Pothole recognition (highlighted in red color) and AR visualization of the corresponding information (in the view of ego1).}
\label{00113_r}
\end{figure}

\begin{figure}[h]
\centering
\includegraphics[width=0.5\textwidth]{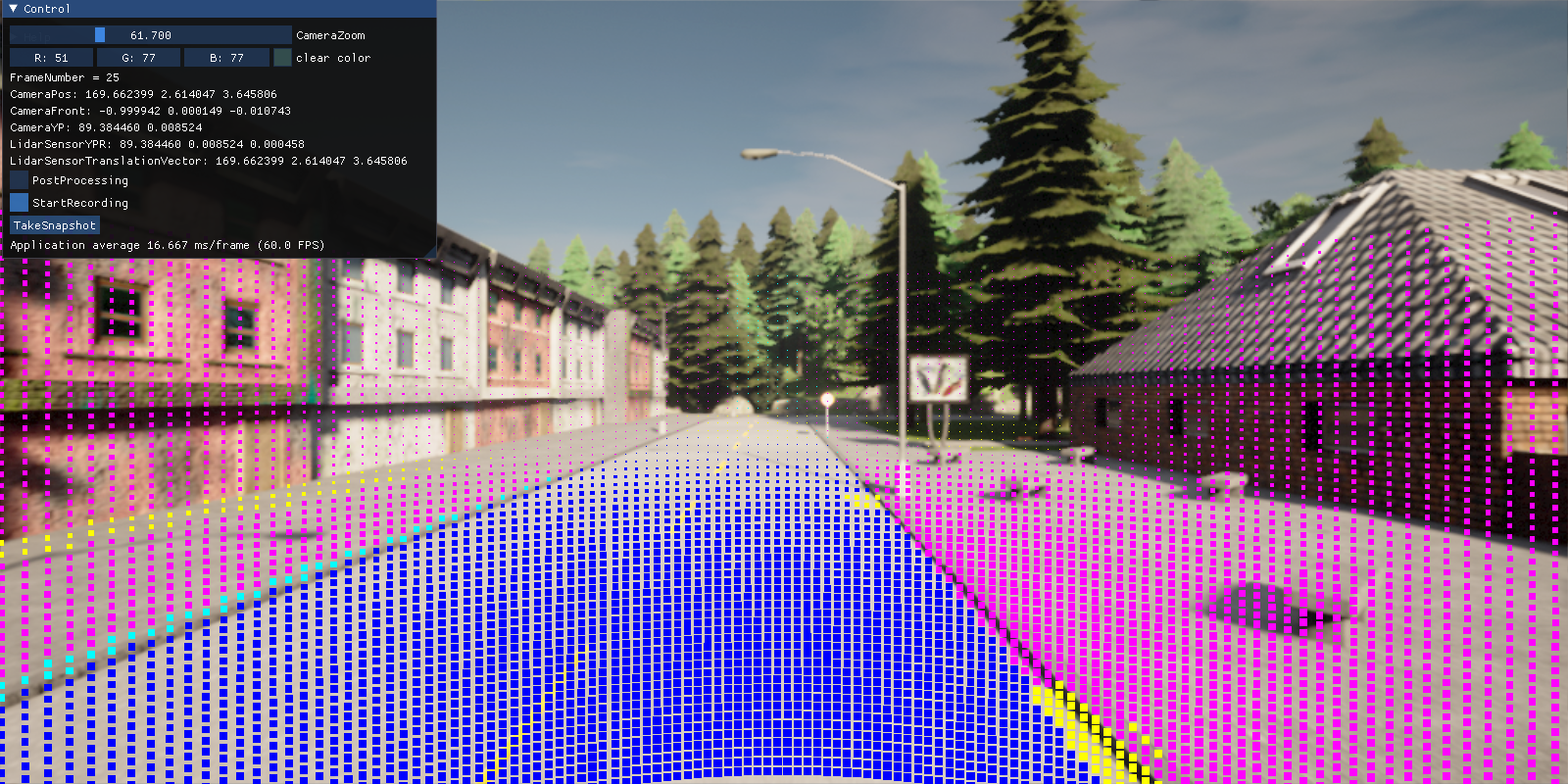}
\caption{AR projection of the point cloud vertices to the scene image that depicts the starting point of view of the ego2 vehicle.}
\label{00025}
\end{figure}

\begin{figure}[h]
\centering
\includegraphics[width=0.5\textwidth]{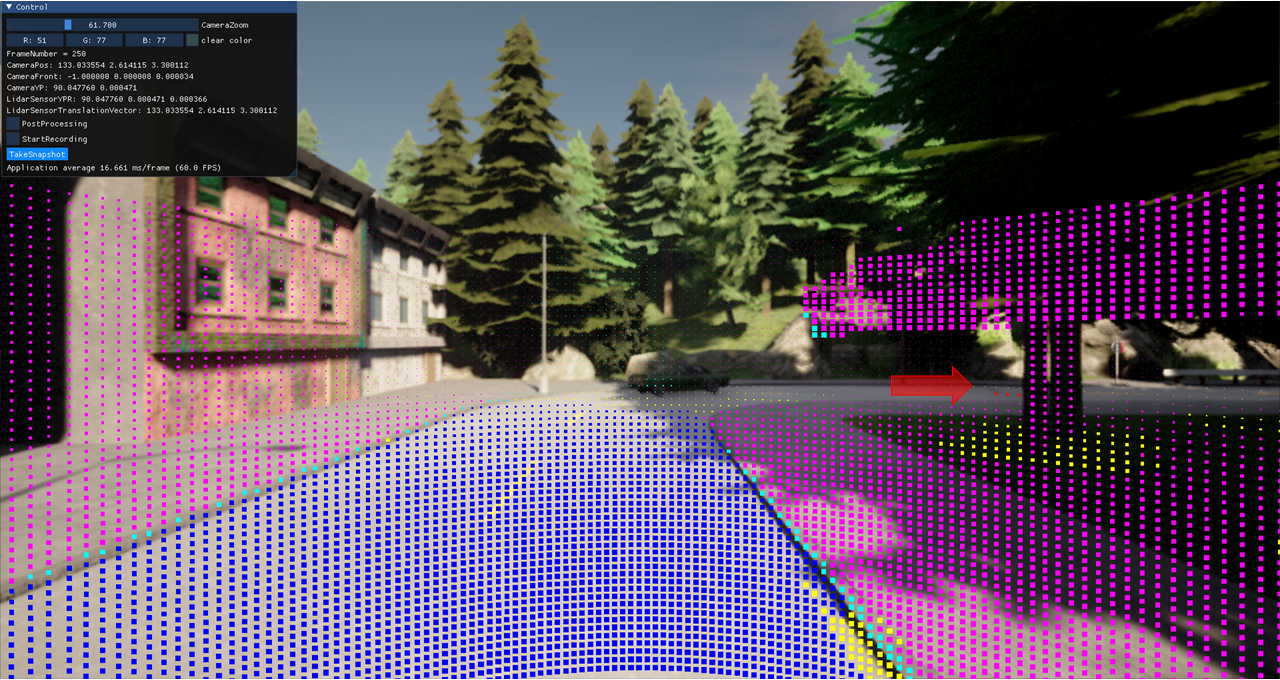}
\caption{Early warning of upcoming pothole to inform ego2.}
\label{00025_r}
\end{figure}

\begin{figure}[h]
\centering
\includegraphics[width=0.5\textwidth]{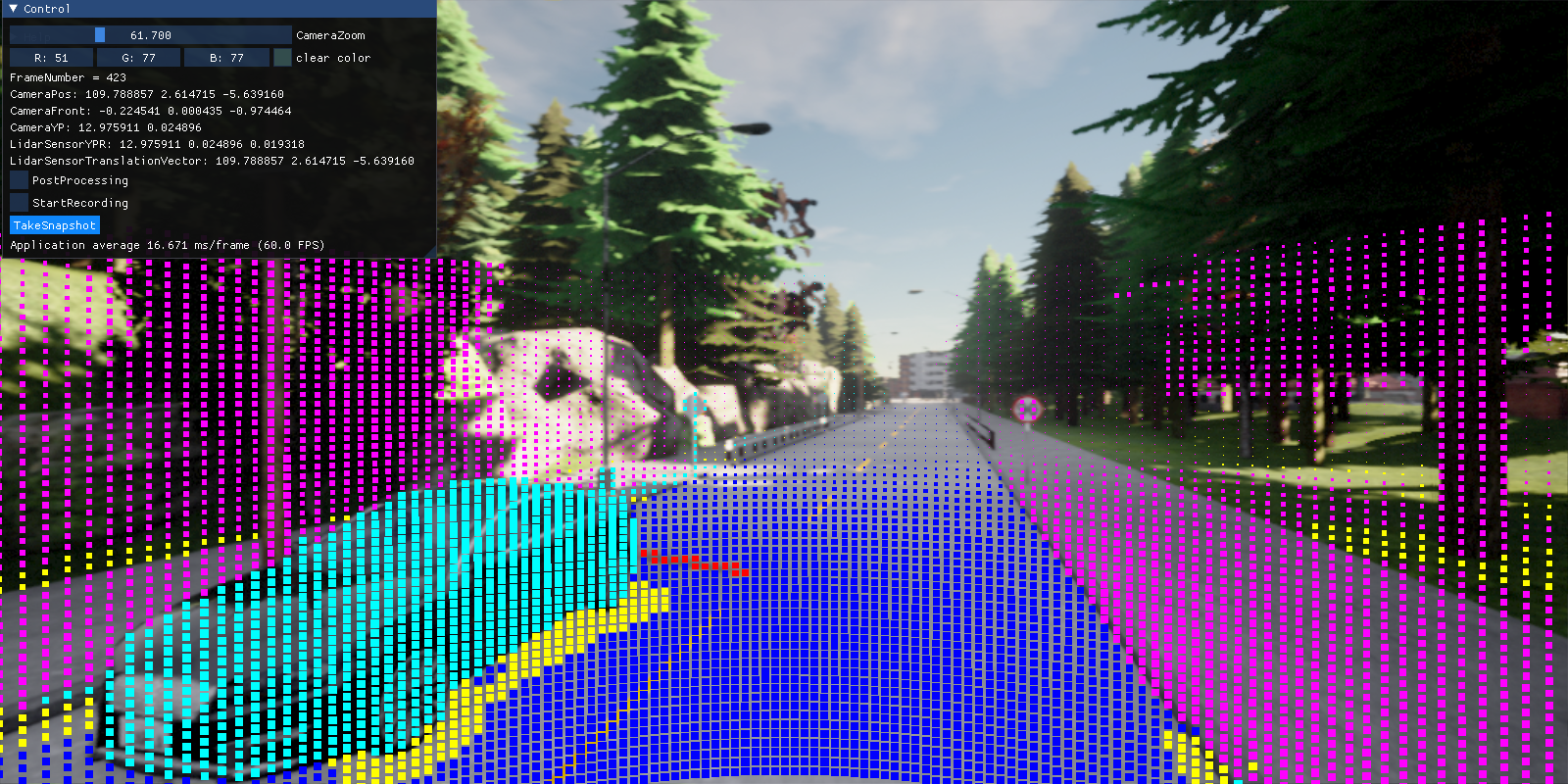}
\caption{Pothole recognition and visualization (in the view of ego2).}
\label{00025_l}
\end{figure}

%%%%%%%%% Dataset subsection %%%%%%%%% 
\input{dataset}

\section{Interfaces and Communication}
\label{sec:visualization_communication}
Context-awareness is a critical factor for successful take-over requests and a lot of effort has been devoted to determining the type of stimulus (e.g. visual, auditory, vibrotactile) \cite{bazilinskyy2017} and the required time-window \cite{white2019, zeeb2015, vivien2015}. %Studies seem to converge on a 10 second time window along with a tie in preference between enhanced visual cues and auditory ones. 
In the case of partial or conditional driving automation, our framework could be used to prepare the driver to quickly take the control of the vehicle, if requested. In order to ensure that the driver is able to swiftly take over the control of the vehicle in an efficient way, we developed a notification system that presents relevant information about the condition of the environment. Our notification system is based on non-intrusive visual cues to prevent tunnel visioning, alerting the driver of potential risks and also directing his/her attention to the objects of interest that sparked the take-over request. In that way, in addition to assisting the human operator during manual driving, the system can, in times of automated driving, trigger the attention of the operator to possible external hazards and preparing him/her to resume control.%, while also focusing his/her attention to highlighting the objects of interest that triggered the request. 
The visualization technique presented in this section is designed as an AR windshield interface, although this is not restrictive, i.e. the method can be implemented in any AR interface. 

\subsection{AR Visualization}
\label{subsec:visualization}
% The visualization targets two different use cases. The first case is for day-to-day cruising which includes utilities and quality of life applications, while the second case concerns the handover scenario where the driver resumes manual control of the vehicle.
%\subsubsection{Utilities}
%\color{red}
%Here, we could potentially write about stuff like google %maps visualization and other QoL improvements.
%\color{black}

The visualization of obstacles is performed by projection. Assuming the position is known for the AR interface and the LiDAR relative to the world, we construct a transformation matrix to map the points of the point cloud from the LiDAR relative coordinate system to the AR interface's coordinate system. The transformation between two different coordinate systems is typically performed by applying serially a scale, a rotation and then a translation transformation. Since both coordinate systems are orthonormal, the scaling can be omitted. Also, by taking advantage of the rigid body nature of the vehicle where the LiDAR and AR interface is located, we also omit the rotation matrix given that, without loss of generality, we can assume that the two coordinate systems are aligned. According to these assumptions, the LiDAR coordinates are transformed into the AR interface's coordinates by a simple translation.

For projecting the points of the point cloud to the AR interface, we assume a simple pinhole camera model. If the AR interface is, for example, an AR windshield, then the windshield represents the image plane and the head of the driver the principal point with coordinates $(x_0, y_0)$. That way, the focal distance $f=(f_x, f_y)$ represents the distance from the driver to the image plane. With the dimensions of the image plane (windshield), and specifically the aspect ratio, known, the frustum is fully defined and the projection can be made from a point in 3D windshield coordinates $(x,y,z)$ to pixels $(u,v)$ on the image plane using the following equation: 
\[
\centering
\begin{pmatrix}
u\\
v
\end{pmatrix}
=
\begin{pmatrix}
1 & 0 & x_0\\
0 & 1 & y\\
0 & 0 & 0
\end{pmatrix}
\begin{pmatrix}
f_x & 0 & 0\\
0 & f_y & 0\\
0 & 0 & 0
\end{pmatrix}
\begin{pmatrix}
x\\
y\\
z
\end{pmatrix}
\]
An undesirable property is the sparsity of the projected pixels attributed to the sparsity of the point cloud. To overcome this limitation, we use an iterative nearest neighbour algorithm on the image space to fill the gaps between projected points. The result of this process is shown in Figs. \ref{00113}-\ref{00025_l}. 

More specifically, Fig.~\ref{00113} and Fig.~\ref{00025} illustrate the segmented point cloud projected to the AR interface of ego1 and ego2 correspondingly. Note that all information is rendered for the sake of completeness. In real-world cases only the necessary information (e.g., arrows or recognised potholes) will be rendered so as to avoid clutter. Fig.~\ref{00113_1} shows the perspective projection of the points to the AR interface and image filling for ego1. In Fig.~\ref{00113_r} and Fig.~\ref{00025_l}, the pothole recognition and visualization is depicted for the vehicles ego1 and ego2, while a warning about an upcoming pothole (retrieved from the database) before reaching the field of view of ego2 is presented in Fig.~\ref{00025_r}.

We would like to clarify here that for evaluation of our methodology and demonstration purposes in the previous figures we project and illustrate in the 2D display device all the information from scene segmentation. However, in real driving scenarios only the most relevant information of the scene (e.g., dangerous objects, potholes) would be highlighted and displayed so as to decrease the amount of any unnecessary information that may bother or confuse the driver.

\subsection{Information Storage and Vehicle Communication Rules}
\label{subsec:communication}

One of the advantages of autonomous vehicles is their ability to communicate with each other forming a cyber-physical system of systems. Many new opportunities arise from the ability of systems to share information, one of which is the transmission of objects or landmarks of interest that were previously observed by an agent, to other agents of the system who could benefit from such information. In particular, our work focuses on information sharing among vehicles about encountered obstacles, such as potholes and bumps, through a centralized server. When a vehicle identifies an unexpected (i.e., unregistered) obstacle, the vehicle sends a request to the server and after further inspection, the new potential obstacle is either discarded or added to the database. Vehicles may also send information regarding already known obstacles when they come across them. Such information includes the Global Positioning System (GPS) location, dimensions and geometrical characteristics in case the obstacle needs updating in the database, e.g. it has increased in size or has been fixed. Through this communication system, a driver can be warned about potential hazards that may not yet be in his field of view or they are obstructed by other objects and thus, increase his performance and decision-making abilities. We should clarify that our work does not focus on communication protocols and defence mechanisms against potential network attacks, but rather defines a solid framework describing the roles of each node and the information flow.

By using the LiDAR-based obstacle detection method, described in section \ref{sec:obstacle_detection}, the vehicle transmits via a communication component to a central server the points belonging to the obstacle, segmented from the point cloud scene. The information is coupled with a timestamp and the GPS location of the vehicle at that instance. The server then transmits to any vehicle in the vicinity of the obstacle, alerting (autonomous vehicles or human operators) about potential hazards from a large distance and thus helping alleviate the inability of the LiDAR sensor to identify obstacles from such a range. In the case of a driver, we also use the AR interface of the vehicle to display, in a non-distracting manner, the location and nature of the potentially upcoming obstacle.

Potholes can change shape over time, most commonly due to deterioration of the surrounding pavement and erosion caused by environmental effects or in the opposite case due to pothole repair. Thus, periodic updates are necessary for the long-term reliability of the pothole visualization component. As there is a need for periodical evaluation of the objects in the server database and update in the case of changes\comm{(potholes being repaired or worsened)}, we assign a shape- and geometry-based descriptor at each obstacle, so that it is characterized by a unique representative signature. Thus, every vehicle encountering the obstacle in a nearby range, calculates the descriptor of the obstacle's area. The new descriptor is then transmitted to the server and is used to confirm whether the information is up-to-date. In the case of a difference in the descriptor's value, an algorithm running in the server decides between keeping the old descriptor, updating it with the new one, or marking the obstacle as removed and deleting the entry from the database. 

\begin{figure}[h]
\centering
\includegraphics[width=0.5\textwidth]{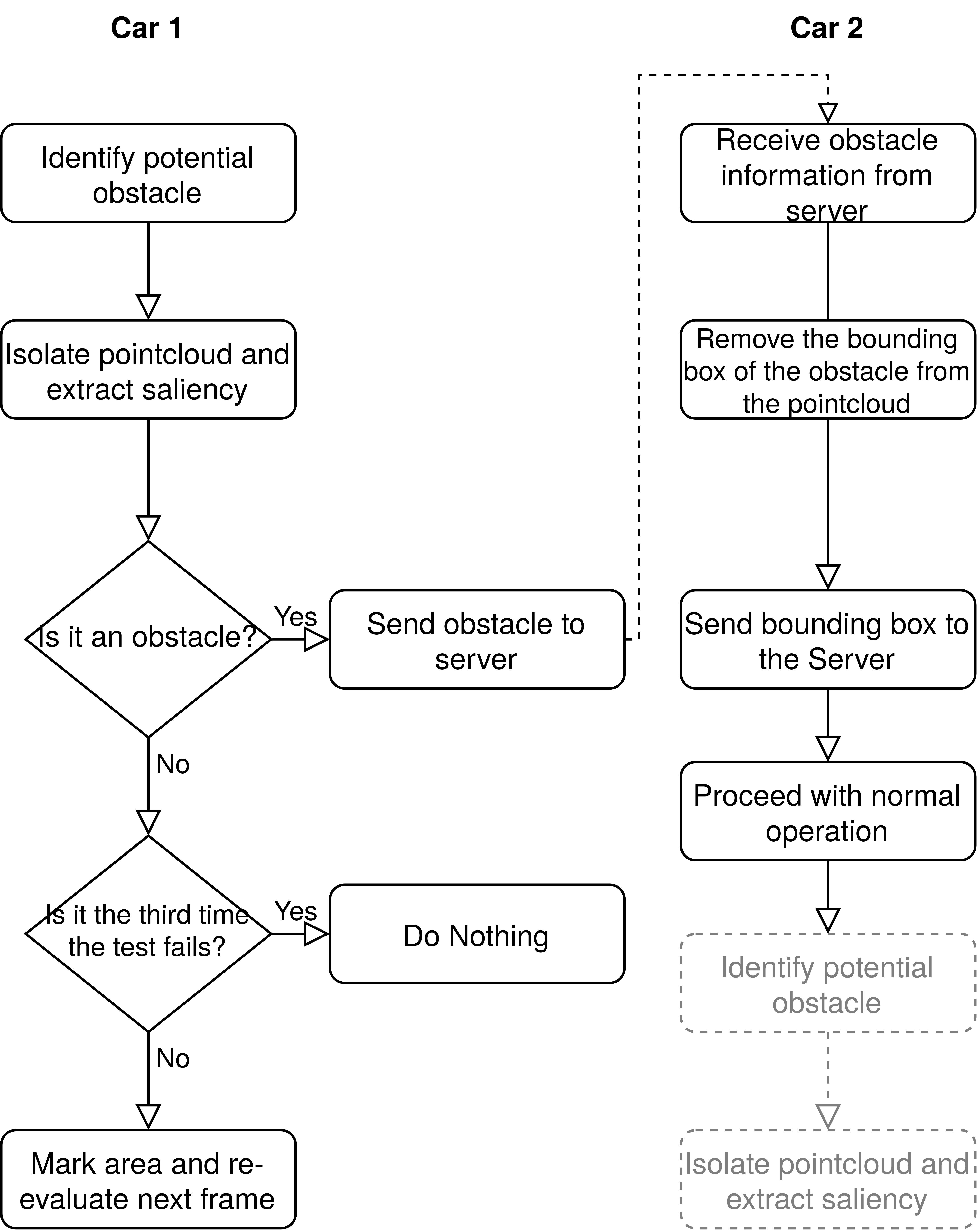}
\caption{Flowchart of communicative vehicles for obstacle sharing.}
\label{flowchart}
\end{figure}

More specifically, we implement a simple system %where a known pothole when seen by another vehicle is checked against the database to determine any possible change and, if needed, update the database.
that (when a new pothole is detected) initiates a database search to retrieve whether the pothole is new or already existed and needs to be updated. Since potholes are static and thus change only in shape, the similarity check is based only on the bounding box of the re-identified pothole. When the overlap of the bounding boxes is less than %When the changing area of the box falls outside 
a threshold, the previous object is replaced by the new one. In our experiments we used a threshold of 15\% reshape in the area in either direction to avoid frequent unnecessary updates, while also retaining the required precision in representation. Similarly, the algorithm checks for significant changes in the bounding box dimensions\comm{(e.g., diagonal lines) and the saliency values of the vertices between the different versions}.
A flowchart showcasing the information update and communication pipeline between two vehicles is shown in Fig.~\ref{flowchart}.

%\begin{comment}
%\begin{figure*}[h]
%\centering
%\includegraphics[width=0.95\textwidth]{img/10784.png}
%\caption{Pothole hidden by the shadow.}
%\end{figure*}

%\begin{figure*}[h]
%\centering
%\includegraphics[width=0.95\textwidth]{img/10832.png}
%\caption{bumps.}
%\end{figure*}
%\end{comment}

\begin{figure*}[h]
\centering
\includegraphics[width=0.95\textwidth]{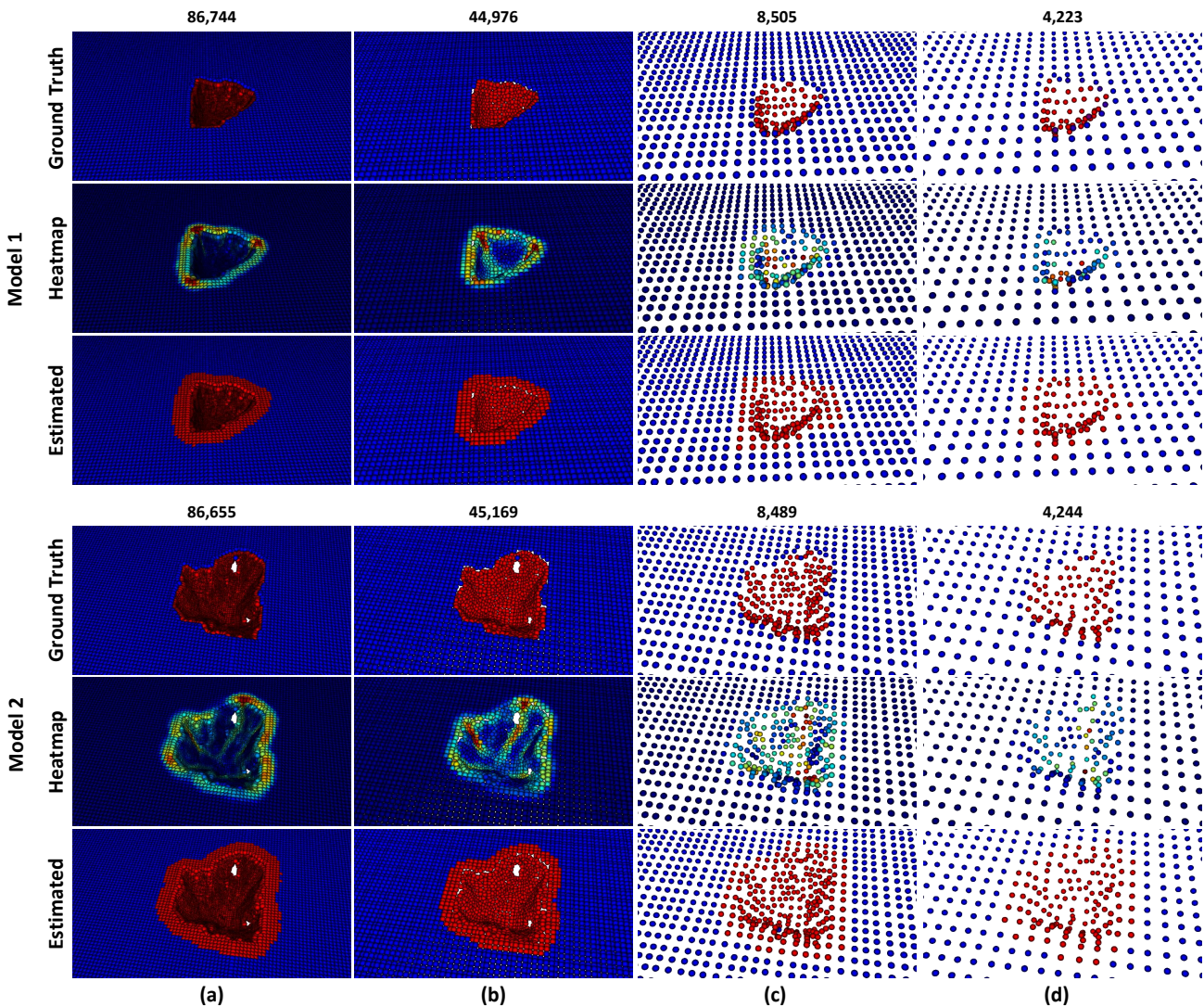}
\caption{Pothole detection in point cloud data of real potholes \cite{fan2020tcyb}. Two dense models are visualized: model1 (rows 1-3) and model2 (rows 4-6). For each model, the three rows illustrate (i) the ground truth, (ii) the heatmap visualizing the saliency map of the pothole and (iii) the estimated point cloud, respectively. The columns show results with decreasing density resolutions (in respect to the original model): (a) original model, \comm{(b) model simplified by $\sim$0.5, (c) model simplified by $\sim$0.1, (d) model simplified by $\sim$0.05.}(b) $\sim50\%$ of the vertices, (c) $\sim10\%$ of the vertices, (d) $\sim5\%$ of the vertices.}
\label{pothole_detection}
\end{figure*}

\section{Experimental Analysis}
\label{sec:experiments}
In this section, we will present and discuss in detail the experimental analysis and will evaluate our proposed framework.

\subsection{Experimental Setup, Datasets and Metrics}
%Experimental setup
The experiments were carried out on an Intel Core i7-4790HQ CPU @ 3.60GHz PC with 16 GB of RAM. The core algorithms are written in Matlab and C++. 
%Datasets
The evaluation of the methodology was performed using (i) synthetic dataset of potholes that we have created  and (ii) 3D point cloud potholes from real datasets with known models (used as ground truth) which have been evaluated by other methods too \cite{fan2020tcyb, fan2019tip, fan2019road, fan2018road}.
%Evaluation metrics

The pothole detection algorithms are compared in terms of the pixel-level (for image-based methods) and point-level (for point clouds) $precision = [TP/(TP + FP)]$, $recall = [TP/(TP+FN)]$, $accuracy = [(TP + TN)/(TP + TN + FP + FN)]$ and $F-score = 2 \cdot [(precision \cdot recall)/(precision + recall)] $, where $TP, \ FP, \ TN, \ FN$, represent the number of True-Positive, False-Positive, True-Negative and False-Negative pixels, respectively. The positive class  includes all vertices belonging to the pothole ($P$) and the negative class all vertices belonging to the road ($R$). The performance metrics can also be expressed as shown in Table \ref{table_eval}, where \textit{Real Pothole} (RP) represents the recall\comm{precision} or in other words the percentage of vertices correctly annotated as pothole, \textit{Real Road} (RR) represents the percentage of vertices correctly annotated as road, \textit{Not real Pothole} (NP) represents the percentage of vertices wrongly annotated as pothole and \textit{Not real Road} (NR) represents the percentage of vertices wrongly annotated as road.

\input{Table_Eval}

%\begin{figure*}[h]
%\centering
%\includegraphics[width=0.95\textwidth]{img/pothole_comparisons1}
%\caption{potholes_comparisons}
%\end{figure*}

\begin{table*}
  \caption{\label{table1} Pothole detection accuracy (in percentage \%) for different density resolutions of the point cloud models.}
\small %\footnotesize %\small
\begin{center}   %original  50%  95%  200% %%I have also results for 70% simplification but the table will be very large
\begin{tabular}{ |c | c | c || c | c  || c | c || c | c | } 
  \hline
 Models &  \multicolumn{2}{c||}{Original} & \multicolumn{2}{c||}{$\sim0.5 \ *$  \text{Original}} & \multicolumn{2}{c||}{$\sim0.1 \ *$ \text{Original}} & \multicolumn{2}{c|}{$\sim0.05 \ *$ \text{Original}} \\ \hline
 \multirow{2}{*}{Model 1} & \text{RP} = 100 & NR = 0 & \text{RP} = 100  & NR = 0  & \text{RP} = 100  & NR = 0  & \text{RP} = 100  & NR = 0  \\ \cline{2-9}
  & $\text{NP}$ = 0.44   & RR = 99.56  & $\text{NP}$ = 0.40  & RR = 99.60  & $\text{NP}$ = 0.68  & RR = 99.32  & $\text{NP}$ = 0.89  & RR = 99.11 \\ 
  \hline  %\hline
   \multirow{2}{*}{Model 2} & \text{RP} = 100  & NR = 0  & \text{RP} = 100  & NR = 0  & \text{RP} = 99.38  & NR = 0.62  & \text{RP} = 100  & NR = 0 \\\cline{2-9}
  & $\text{NP}$ = 0.65   & RR = 99.35  & $\text{NP}$ = 0.58  & RR = 99.42  & $\text{NP}$ = 0.98  & RR = 99.02  & $\text{NP}$ = 1.22  & RR = 98.78 \\ 
  \hline
   \multirow{2}{*}{Model 3} & \text{RP} = 100  & NR = 0  & \text{RP} = 100  & NR = 0  & \text{RP} = 100  & NR = 0  & \text{RP} = 100  & NR = 0 \\\cline{2-9}
  & $\text{NP}$ = 0.41   & RR = 99.59  & $\text{NP}$ = 0.36  & RR = 99.64  & $\text{NP}$ = 0.63  & RR = 99.37  & $\text{NP}$ = 0.74  & RR = 99.26 \\ 
  \hline
   \multirow{2}{*}{Model 4} & \text{RP} = 99.91  & NR = 0.08  & \text{RP} = 100  & NR = 0  & \text{RP} = 100  & NR = 0  & \text{RP} = 100  & NR = 0 \\\cline{2-9}
  & $\text{NP}$ = 0.45   & RR = 99.55  & $\text{NP}$ = 0.41  & RR = 99.59  & $\text{NP}$ = 0.76  & RR = 99.24  & $\text{NP}$ = 0.93  & RR = 99.07 \\ 
  \hline
   \multirow{2}{*}{Model 5} & \text{RP} = 100  & NR = 0  & \text{RP} = 100  & NR = 0  & \text{RP} = 99.36  & NR = 0.64  & \text{RP} = 100  & NR = 0 \\\cline{2-9}
  & $\text{NP}$ = 0.62   & RR = 99.38  & $\text{NP}$ = 0.56  & RR = 99.44  & $\text{NP}$ = 0.86  & RR = 99.14  & $\text{NP}$ = 1.25  & RR = 98.75 \\ 
  \hline
   \multirow{2}{*}{Model 6} & \text{RP} = 100  & NR = 0   & \text{RP} = 99.57  & NR = 0.43  & \text{RP} = 100  & NR = 0  & \text{RP} = 100  & NR = 0 \\\cline{2-9}
  & $\text{NP}$ = 0.52   & RR = 99.48   & $\text{NP}$ = 0.46  & RR = 99.54  & $\text{NP}$ = 0.78  & RR = 99.22  & $\text{NP}$ = 1.22  & RR =  98.78 \\
  \hline
   \multirow{2}{*}{Model 7} & \text{RP} = 99.92  & NR = 0.08   & \text{RP} = 100  & NR = 0  & \text{RP} = 99.15  & NR = 0.85  & \text{RP} = 100  & NR = 0 \\\cline{2-9}
  & $\text{NP}$ = 0.47   & RR = 99.53  & $\text{NP}$ = 0.44  & RR = 99.56  & $\text{NP}$ = 0.80  & RR = 99.20  & $\text{NP}$ = 1.10  & RR = 98.90 \\ 
  \hline
   \multirow{2}{*}{Model 8} & \text{RP} = 100  & NR = 0  & \text{RP} = 100  & NR = 0  & \text{RP} = 100  & NR = 0  & \text{RP} = 100  & NR = 0 \\\cline{2-9}
  & $\text{NP}$ = 0.48   & RR = 99.52  & $\text{NP}$ = 0.44  & RR = 99.56  & $\text{NP}$ = 0.73  & RR = 99.27  & $\text{NP}$ = 1.06  & RR = 98.94 \\ 
  \hline
   \multirow{2}{*}{Model 9} & \text{RP} = 100  & NR = 0  & \text{RP} = 99.13  & NR = 0.87  & \text{RP} = 100  & NR = 0  & \text{RP} = 100  & NR = 0 \\\cline{2-9}
  & $\text{NP}$ = 0.45   & RR = 99.55  & $\text{NP}$ = 0.37  & RR = 99.63  & $\text{NP}$ = 0.65  & RR = 99.35  & $\text{NP}$ = 0.96  & RR = 99.04 \\ 
  \hline
   \multirow{2}{*}{Model 10} & \text{RP} = 99.92  & NR = 0.08  & \text{RP} = 100  & NR = 0  & \text{RP} = 100  & NR = 0  & \text{RP} = 100  & NR = 0 \\\cline{2-9}
  & $\text{NP}$ = 0.42   & RR = 99.58  & $\text{NP}$ = 0.38  & RR = 99.62  & $\text{NP}$ = 0.68  & RR = 99.32  & $\text{NP}$ = 0.84  & RR = 99.16 \\ 
  \hline
   \multirow{2}{*}{Model 11} & \text{RP} = 100  & NR = 0  & \text{RP} = 100  & NR = 0  & \text{RP} = 100  & NR = 0  & \text{RP} = 98.56  & NR = 1.44 \\\cline{2-9}
  & $\text{NP}$ = 0.59   & RR =99.41  & $\text{NP}$ = 0.51  & RR = 99.49  & $\text{NP}$ = 0.95  & RR = 99.05  & $\text{NP}$ = 1.03  & RR = 98.97 \\ 
  \hline
   \multirow{2}{*}{Model 12} & \text{RP} = 100  & NR = 0  & \text{RP} = 99.36  &  NR = 0.64  & \text{RP} = 100  & NR = 0  & \text{RP} = 100  & NR = 0 \\\cline{2-9}
  & $\text{NP}$ = 0.44   & RR = 99.56  & $\text{NP}$ =0.36  & RR = 99.64  & $\text{NP}$ = 0.64  & RR = 99.36  & $\text{NP}$ = 0.81  & RR = 99.19 \\ 
  \hline
   \multirow{2}{*}{Model 13} & \text{RP} = 99.96  & NR = 0.04  & \text{RP} = 100  & NR = 0  & \text{RP} = 100  & NR = 0  & \text{RP} = 100  & NR = 0 \\\cline{2-9}
  & $\text{NP}$ = 0.42   & RR = 99.58  & $\text{NP}$ = 0.38  & RR = 99.62  & $\text{NP}$ = 0.65  & RR = 99.35  & $\text{NP}$ = 0.79  & RR = 99.21 \\ 
  \hline
   \multirow{2}{*}{Average} & \text{RP} = 99.98  & NR = 0.02  & \text{RP} = 99.85  & NR = 0.15  & \text{RP} = 99.83  & NR =  0.17  & \text{RP} = 99.88  & NR = 0.11 \\\cline{2-9}
  & $\text{NP}$ = 0.49   & RR = 99.51  & $\text{NP}$ = 0.43  & RR = 99.57  & $\text{NP}$ = 0.75  & RR = 99.25  & $\text{NP}$ = 0.99  & RR = 99.01\%\\ 
  \hline
\end{tabular}
\end{center}
\end{table*}

\begin{figure*}%[H]
\centering
\includegraphics[width=0.99\textwidth]{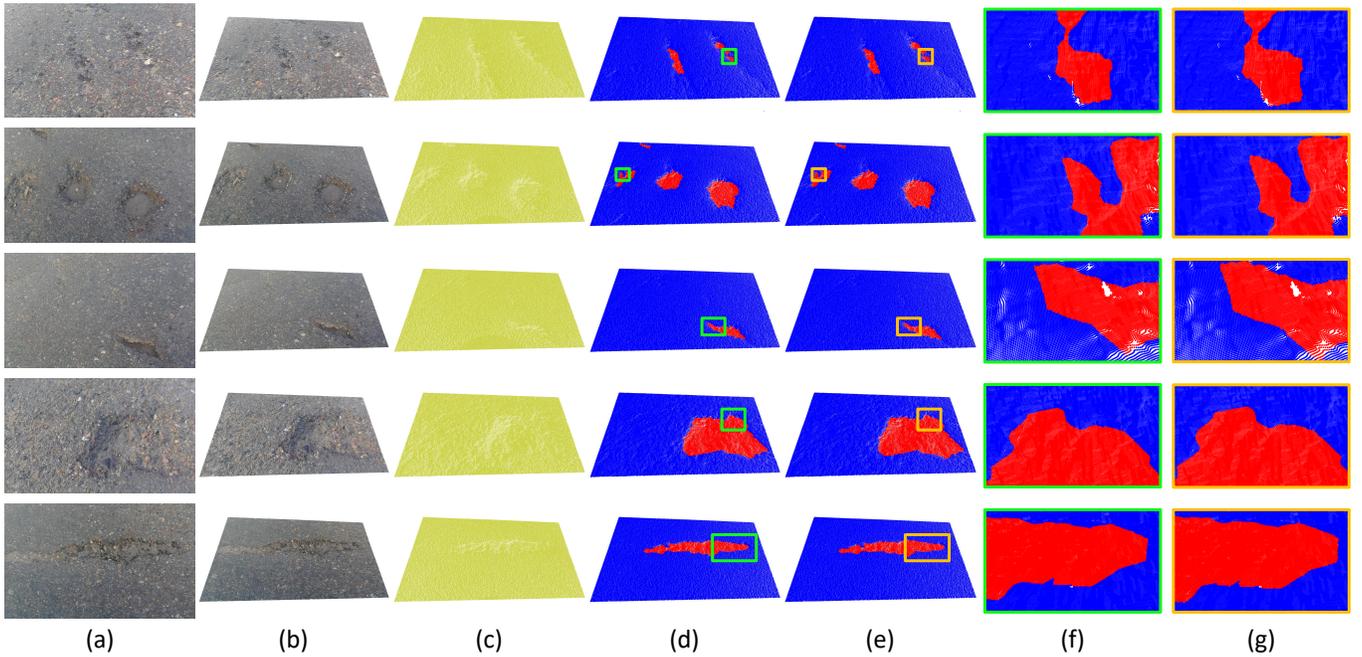}
\caption{Pothole detection on real data \cite{fan2019tip}. (a) RGB images of potholes, (b) corresponding point cloud of potholes with texture, (c) point cloud of potholes, (d) ground truth binary mask of potholes, (e) estimated binary mask of potholes, (f) enlarged details of the ground truth point cloud, (g) enlarged details of the estimated point cloud.}
\label{pothole_detection2}
\end{figure*}

\begin{table*}[hbt]
\normalsize
\caption{\label{table2}Comparison of the pothole detection accuracy among different state-of-the-art approaches.}
\normalsize
\begin{center}   %original  50%  95%  200% %%I have also results for 70% simplification but the table will be very large
\begin{tabular}{ |c | c | c | c | c  | c | c | c | c | } 
  \hline
 Dataset &  Method & Correct Detection & Incorrect & Misdetection & Recall & Precision & Accuracy & F-score \\ \hline
 \multirow{5}{*}{ 1} & 1 \cite{6853659} & 11 & 11 & 0 & 0.520 & 0.543 & 0.989 & 0.531 \\ %\cline{2-9}
  & 2 \cite{c45f903c61ed4409bb32f8fc4f0220f3} & 22 & 0 & 0 & 0.462 & 0.998 & 0.994 & 0.632 \\ 
  & 3 \cite{fan2019tip} & 22 & 0 & 0 & 0.499 & 0.987 & 0.994 & 0.663 \\
  & 4 \cite{fan2020tcyb} & 21 & 1 & 0 & 0.701 & 0.964 & 0.995 & 0.811 \\
  & our & 22 & 0 & 0 & 0.853
 & 0.993 & 0.991 & 0.918 \\
  \hline  %\hline
   \multirow{5}{*}{2}  & 1 \cite{6853659} & 42 & 10 & 0 & 0.975 & 0.971 & 0.999 & 0.973 \\
   & 2 \cite{c45f903c61ed4409bb32f8fc4f0220f3} & 40 & 8 & 4 & 0.874 & 0.991 & 0.997 & 0.929 \\
   & 3 \cite{fan2019tip} & 51 & 1 & 0 & 0.980 & 0.980 & 0.999 & 0.980 \\
   & 4 \cite{fan2020tcyb} & 52 & 0 & 0 & 0.950 & 0.883 & 0.992 & 0.915 \\
   & our & $40^2$ & 0 & 0 & 0.909 & 0.996 & 0.992 & 0.951 \\
  \hline
   \multirow{5}{*}{3} & 1 \cite{6853659} & 5 & 0 & 0 & 0.612 & 0.771 & 0.995 & 0.683 \\
   & 2 \cite{c45f903c61ed4409bb32f8fc4f0220f3} & 5 & 0 & 0 & 0.534 & 0.992 & 0.996 & 0.694 \\
   & 3 \cite{fan2019tip} & 5 & 0 & 0 & 0.582 & 0.983 & 0.996 & 0.731 \\
   & 4 \cite{fan2020tcyb} & 5 & 0 & 0 & 0.702 & 0.996 & 0.996 & 0.823 \\
   & our & 5 & 0 & 0 & 0.953 & 0.984 & 0.996 & 0.969 \\
   \hline
    \multirow{5}{*}{Total} & 1 \cite{6853659} & 58 & 21 & 0 & 0.800 & 0.822 & 0.994 & 0.800 \\
   & 2 \cite{c45f903c61ed4409bb32f8fc4f0220f3} & 67 & 8 & 4 & 0.695 & 0.992 & 0.995 & 0.817 \\
   & 3 \cite{fan2019tip} & 78 & 1 & 0 & 0.771 & 0.982 & 0.996 & 0.864 \\
   & 4 \cite{fan2020tcyb} & 78 & 1 & 0 & 0.890 & 0.898 & 0.996 & 0.894 \\
   & our & 67 & 0 & 0 & 0.899 & 0.994 & 0.992 & 0.945 \\ 
  \hline
 % \hline
\end{tabular}
\end{center}
\footnotesize{$^2$ Only 40 of the refereed (52) models were founded online.}
\end{table*}

\normalsize

\subsection{Results}

For the evaluation of our method, two public available datasets \cite{fan2020tcyb, fan2019tip} were utilized providing point clouds of real potholes. Fig.~\ref{pothole_detection} visualizes results of our pothole detection method for the dataset created by real potholes \cite{fan2020tcyb} under different density resolutions (\ref{pothole_detection} (a)-(d)). Points in red represent the vertices belonging to the pothole, while points in blue represent vertices belonging to the road, both for the ground truth and the estimated point clouds. 
Two dense models (\ref{pothole_detection} (a)) are utilized as presented in rows 1-3 and 4-6, respectively. To investigate the performance of our approach in more realistic conditions, we increasingly downsampled the original point cloud (\ref{pothole_detection} (b)-(c)) to evaluate the robustness of detection of our algorithm. The corresponding number of vertices for the two models (original and downsampled) are shown above each model, respectively. The heatmap (rows 2 and 5) illustrates the geometric and spectral saliency per vertex (as estimated from Eq. \ref{combined_saliency}). Higher salient values are depicted with deep red color while lower salient values with deep blue.

Due to the sensitive nature of the specific application involving safety of drivers (via information visualization for situational awareness), we prefer our algorithm to provide a small percentage of $\text{NR}$ than having even a small value of $\text{NP}$ (please refer to Table \ref{tableI}). To wrongly identify as a pothole a small area of the road around an actual pothole is not as critical in our application as the opposite, namely to fail to present or partially present a potentially dangerous object (e.g., pothole, ramp).

The detailed results with all evaluation metrics are shown in Table \ref{table1} for each of the thirteen 3D models of the point cloud dataset, and under different point cloud density resolutions. The results of this table show that our method is robust even for very low point cloud density. This is an important observation, since the output of the LiDAR device has a low density resolution pattern. 

Fig.~\ref{pothole_detection2} visualizes some examples of the pothole detection algorithm applied in an other dataset \cite{fan2019tip}. The first column of this figure illustrates the RGB image presenting real road potholes. In the second column (Fig.~\ref{pothole_detection2}-(b)), the corresponding point cloud with the relative texture is presented. The geometry represented by the 3D coordinates of the point cloud (without any color information) is presented in Fig.  \ref{pothole_detection2}-(c). Fig.~\ref{pothole_detection2}-(d) shows the ground truth vertices (in red) representing the pothole and Fig.~\ref{pothole_detection2}-(e) presents our pothole estimation result. Figs. \ref{pothole_detection2}-(f) $\&$ (g) just present enlarged details of Figs. \ref{pothole_detection2}-(d) $\&$ (e), respectively, for easier visual comparison.

Table \ref{table2} provides a qualitatively comparison of our method versus other approaches of the literature. However, it should be mentioned that the results are not directly comparable because the other methods use only the visual information of the RGB images, while our method uses only the geometrical information of the corresponding point cloud. 

\section{Conclusions}
\label{sec:conclusions}
In this paper we presented a methodology for identification of road obstacles and their AR-based visualization targeting both the driver of the \textit{ego} vehicle%that his/her sensor's system received and proceed the relative point cloud information, as well as 
and other drivers in a spatial vicinity whose LiDAR device has not captured  the obstacle information yet. AR-enabled technologies (beyond current AR headsets) are expected to be utilized in the near future for providing guidance to the drivers \cite{8699235, 8814237, 9756811}, increasing their situational awareness, and facilitating cooperation with other vehicles and road users (e.g., pedestrians, bicycles). %Since the automotive industry is adopting AR interfaces in the form of windshields. 
The main purpose of the proposed system is to be capable to provide in real-time information to the drivers of autonomous and connected vehicles in cooperative driving situations, in order to increase their situational awareness.

Main emphasis was placed to the detection of potholes rather than protruding obstacles, because missing parts of the road present particular challenges that have not been handled efficiently by the available methods so far \cite{9676673}. Our method is based on the analysis of point clouds which is challenged by the lack of benchmark datasets obtained from LiDAR devices. To overcome this problem, we created our own synthetic dataset and added it to the maps of the CARLA simulator, thereby creating realistic driving environments. The comparison of our method with other state-of-the-art approaches, regarding the accuracy of pothole detection in real datasets, has shown its effectiveness providing very promising outcomes. 

Our future plans include the visualization of additional information that can facilitate the increase of driver's situational awareness (e.g., road boundaries), and the analysis of user preferences, e.g., via questionnaires, of the AR visualization system when driving (through a steering wheel chair) in the simulated environment of the CARLA simulator.

\section*{Appendix}
\subsection{Robust Principal Component Analysis (RPCA)}
\label{appendix:RPCA}
\input{RPCA.tex}

%\section*{Acknowledgment}

% Can use something like this to put references on a page
% by themselves when using endfloat and the captionsoff option.
\ifCLASSOPTIONcaptionsoff
  \newpage
\fi

\bibliographystyle{IEEEtran}
\bibliography{mybibfile}

% if you will not have a photo at all:
%\begin{IEEEbiographynophoto}{John Doe}
%Biography text here.
%\end{IEEEbiographynophoto}

%\begin{IEEEbiographynophoto}{Jane Doe}
%Biography text here.
%\end{IEEEbiographynophoto}

\end{document}

%% file: previous_work.tex
\section{Previous Work}
\label{sec:previous_work}
In the following we provide an overview of methodologies tackling the main challenges of the presented approach on (i) obstacle detection, (ii) cooperative driving and (iii) AR infotainment systems.

\subsubsection{Obstacle Detection}
%With the rapid development of self-driving cars \cite{9409701, 9637501}, several challenges have risen to ensure a safe, stable and predictable cruise. 
A major element that adds unpredictability in path planning for self-driving cars are obstacles in the road. Obstacles can appear in the form of objects beyond the surface of the road, or cracks and holes in paved areas. There has been major work on obstacle detection, raging from real-time implementations \cite{amita2020}, to offline schemes that act as automated informants to the authorities responsible for maintenance \cite{schiopu2016}, or as efficient unsupervised techniques for pothole detection \cite{akagic2017}. Most of the existing works implement a broad spectrum of computer vision and/or machine learning techniques to analyze imaging information \cite{amita2020}. 
The methods differ mainly on the utilized features and classifiers for obstacle representation and recognition. In respect to performance, a direct comparison of methods is not feasible because most works are evaluated on their own (simulated) data. In fact, there is lack or restricted access to a common benchmark dataset with potholes and obstacles, that can be used for comparison. 
%One problem with presenting such methods is that a unified dataset of potholes and obstacles does not exist and thus many researchers had to create their own. As such, comparison between different methods is not feasible, except in the unlikely case that the same dataset is used.

Waqa \etal \cite{waqas2017} used superpixel segmentation to partition the image into superpixels based on the entropy rate, and then applied Support Vector Machines (SVM) to estimate the probability of each superpixel being the part of some object based on textural features (namely histogram of oriented gradients, co-occurrence matrix, intensity histogram and mean intensity). %For each superpixel four features are computed, namely the histogram of oriented gradients, co-occurence matrix, intensity histogram and mean intensity. Due to the fact that the surface an object can span more than one superpixel, merging is performed based on the surrounding extracted features. For the classification of each superpixel, they train different SVM classifiers for each feature. 
For final object label inference, merging of the superpixels is performed using conditional random fields to account for neighborhood similarity. The drawbacks of this method are it's dependency only on texture information and more specifically the inability to distinguish between a shadow and a hole easily, leading to potential false positives.

Another image-based method that takes advantage of the texture characteristics of potholes is the work of Kanza \etal \cite{kanza2016}. Here, the histogram of oriented gradients (HOG) is extracted from the grayscale image and coupled with a Naive Bayes classifier. If the probability calculated by the classifier is high enough, then the pothole localization is performed using graph-based segmentation and normalized cuts. This method presents very encouraging results for the examined dataset that simulates a variety of cases and conditions, including changes in illumination and potholes filled with water.

Yifan \etal \cite{yifan2018} take a different direction and use Unmanned Aerial Vehicle (UAV) for pothole detection in the suburb of Shihenzi City. The aerial images are segmented and the segmented parts are used to extract features, including the mean, standard deviation, area, length/width ratio, elliptic fit, roundness, contrast, dissimilarity, homogeneity and correlation. Segmentation is performed using a multiresolution segmentation algorithm that is integrated into the eCognition Developer software (a development environment for object-based analysis of geospatial data). For classification, the SVM, Artificial Neural Networks (ANN), and Random Forest (RF) classifiers are compared, each with different combinations of features while also taking the computational time into consideration. The authors conclude that spatial features (texture and geometry) coupled with RF are the most effective, although this method is very sensitive to UAV image resolution, weather and lighting conditions.

Other methods focus on road cracks detection from high resolution cameras on smartphones. Since such data are more easily available, those methods can bypass the extraction of hand-crafted features and utilize deep architectures, such as convolutional neural networks \cite{maeda2018, mei2020, yang2019, shi2016}. However, in the case of dense traffic situations and poor lighting conditions, techniques utilizing images from smartphone camera are less effective.

In contrast to computer vision techniques which exploit texture information from images, 3D point cloud processing techniques exploit the object's geometrical properties \cite{chen2017, shuo2017, bosurgi2022}. %, LiDAR-based detection focuses on the object's geometrical properties. 
Bosurgi \etal \cite{bosurgi2022} identify potholes in road sections by estimating area, perimeter and depth information from 3D data of pavement surfaces.
Chen \etal \cite{chen2017} propose a framework for obstacle detection using \comm{the angles $\phi$ and $\theta$, which are} the pitch and rotation angles of a LiDAR sensor to create a 2D image-like plane where the unordered set of points (from the point cloud) are projected. \comm{The rows of this image, called LiDAR-imagery, as well as each line corresponding to a scan line of the sensor, are used to create a 2D histogram.} From this ``LiDAR-imagery" a 2D histogram is extracted and used to find the road plane. If an adequate part of the road, in front of the vehicle is flat, those points form a straight line in the histogram representation, and anything above the line can be classified as a positive obstacle (points higher than the road plane), while points below the line as a negative obstacle (points lower than the road plane). %They also propose a technique to detect potholes filled with water, one of the weaknesses of the LiDAR sensor since, due to refraction and reflection, water appears as a black hole in the LiDAR-imagery. They purport that they can take advantage of this fact and scan the image for large areas of missing data to detect water bodies. 
Moreover, since water bodies cannot be detected by LiDAR due to refraction and reflection, the authors propose a technique to detect potholes filled with water by scanning the image for large areas of missing data.

Shuo \etal \cite{shuo2017} improves the aforementioned method by projecting the points on the camera plane and interpolating the depth values of the projected points to receive a depth image. They use both horizontal and vertical histograms to coarsely detect the road area and refine it respectively. Although they state their method as sensor fusion between the monocular camera and LiDAR, they do not utilize the color values of the camera images. Both works \cite{chen2017} and \cite{shuo2017} use the KITTI dataset as a benchmark and achieve great results, comparable to machine learning methods.

Other techniques for pothole detection may include laser scanning, ground penetrating radar, ultrasonic sensor, as well as multi-sensor fusion, especially concerning fusion with imaging information. An extensive review of such techniques falls beyond the scope of this article. However, an interested reader may be referred to the survey in \cite{ortiz2020}.

\subsubsection{Point cloud saliency}
One of main challenges in techniques utilizing point clouds is the inherent noise and the increased computational cost due to the unordered data structure of point clouds. To address such challenges, saliency map extraction has been proposed as a powerful step in point cloud processing to reduce noise and data dimensionality, leading to more robust solutions and computational efficiency \cite{arvanitis2019} \cite{arvanitis2021}. Yet, the use of local saliency in pothole detection has not been sufficiently examined. Saliency maps were constructed from point clouds obtained from Mobile Laser Scanning (MLS) in \cite{ma2022} for road crack detection. MLS point clouds contain spatial information (i.e., Euclidean coordinates) and intensity information, and thus the extracted features could leverage both height and intensity information. Feature saliency was estimated by calculating the distances from the normal of each point to the principal normal of the input point clouds. 
In a similar setting, Wang \etal \cite{wang2015} extracted saliency maps in MLS point clouds by projecting the distance of each point’s normal vector to the point cloud’s dominant normal vector into a hyperbolic tangent function space.  

\subsubsection{Cooperative driving}
%\begin{comment}
While significant advances have been made for single-agent perception, many applications require multiple sensing agents and cross-agent communication for more accurate results. Objects, captured by the single-agent's sensor devices, may be heavily occluded or far away from the sensors' view, resulting in sparse observations. Nevertheless, failing to detect and predict the accurate position or moving intention of these occluded or ``hard-to-see" objects might have harmful consequences in safety-critical situations, and especially if the reaction time is very narrow \cite{10.1007/978-3-030-58536-5_36}. The development of multi-agent solutions can lead to collaborative perception and, through information sharing, may improve the driving performance and experiences, providing endless possibilities for safe driving. 

Recently, cooperative autonomous driving has been considered as a possible solution to improve the performance and safety of autonomous vehicles \cite{cooperative2020}. Cooperative perception for 3D object detection can be performed via early or late fusion of information, i.e.,  combination of multiple sensing points of view or fusion of object detection results, respectively.Both fusion approaches can extend the perception of the sensing system, however, only the early fusion approach can actually exploit complementary information. 
A major challenge that arises regarding cooperative perception is how to effectively merge sensors' data received from different vehicles to obtain a precise and comprehensive perception outcome. Additionally, despite the attention that cooperative driving has attracted recently, the absence of a suitable open dataset for benchmarking algorithms has made it difficult to develop and assess cooperative perception technologies. 

Xu \etal \cite{2109.07644} presented the first open dataset and used it to benchmark fusion strategies for V2V (vehicle-to-vehicle) perception. They also plan to extend the dataset with more tasks as well as sensor suites and investigate more multimodal sensor fusion methods in the V2V and V2I (vehicle-to-infrastructure) settings. Arnold \etal \cite{cooperative2020} proposed a system that produces a perception of complex road segments (e.g., complex T-junctions and roundabouts) using a network of roadside infrastructure sensors with fixed positions.
Chen \etal \cite{8885377} studied the raw-data level cooperative perception for enhancing the detection ability of self-driving systems. They fuse the sensor data collected from different positions and angles of connected vehicles, relying on LiDAR 3D point clouds.
Liu \etal \cite{9156848} addressed the collaborative perception problem, where one agent is required to perform a perception task and can communicate and share information with other agents on the same task.

Chen \etal \cite{10.1145/3318216.3363300} proposed a point cloud feature-based cooperative perception framework for connected autonomous vehicles to increase object detection precision. The features are selected to be rich enough %The used feature data are selected to be sufficient 
for the training process, and at the same time have an intrinsically small size to achieve real-time edge computing. Guo \etal \cite{9330564} proposed a cooperative fusion method to combine spatial feature maps for achieving a higher 3D object detection performance. Yuan \etal \cite{9682601} proposed a 3D keypoints feature fusion scheme for cooperative driving detection to remedy the problem of low bounding box localization accuracy. 
Fang \etal \cite{9775023} presented an iterated split covariance intersection filter-based cooperative localization strategy with a decentralized framework. In addition, they adopted a point cloud registration method to obtain the\comm{cooperative} relative pose estimation using mutually shared information from neighbour vehicles.
Kim and Liu \cite{7518716} presented the concept of cooperative autonomous driving using mirror neuron-inspired intention awareness and cooperative perception, providing information on the upcoming traffic situations ahead, even beyond line-of-sight and field-of-view.
%\end{comment}

\subsubsection{Situational awareness and AR infotainment}
In the case of semi-autonomous vehicles, where the operator/driver may be asked to take manual control of the car at any moment, it is of great importance \cite{7934009} to implement notification paradigms that direct the operator's, possibly reduced, attention to the event that triggered the take-over request \cite{9261134, 9447801}. Recently, the automotive industry started to invest funds and efforts into AR technology and its integration with In-Vehicle Information Systems (IVIS) for intuitive and non-intrusive information display to the driver.

The design of AR in-vehicle systems for infotainment is a challenging task. Rao \etal \cite{rao2014} performed an analysis of design methods on different use cases aiming to identify the difficulties in implementation aspects. Despite the vast amount of requirements for these systems to work reliably, such as latency, bandwidth, weather conditions etc, they concluded that the integration of augmented reality in vehicles will help drivers navigate their environment better, and thus will be more widely adopted. %Their main idea is that this technology will help drivers navigate their environment better, by presenting important information in an intuitive and non-intrusive way. 
 
While IVIS existing in many modern vehicles with touch Liquid-Crystal Display (LCD) displays and voice commands may seem to offer most of the utilities of an AR infotainment system, they may actually be distracting to the driver. David \etal \cite{strayer2019} showed in a recent study that some IVIS require a high cognitive demand or complex command sequences to be handled, and this can in turn lower the awareness of the operator. This is perpetuated by the fact that most IVIS are placed on the dashboard and usually demand their operation to avert (even momentarily) the driver's gaze from the road. In contrast, AR HUDs perform information rendering on top of the environment and thus the driver does not need to share focus in multiple locations.   

The distraction potential of AR HUDs was assessed by Kim \etal \cite{kim2019}. An AR-enabled windshield was used in a simulated environment with a real-life driving video feed to test various methods of pedestrian visualization. The gaze behavior and cognitive processes were measured and it was found that the visual and cognitive distraction potential of AR depends on the perceptual forms of graphical elements presented on the displays. Specifically, in some cases visualizations, e.g., in the form of a ``virtual transparent shadow" indicating the pedestrian’s anticipated path, improved the driver's attention without degrading awareness of other objects or scene elements. On the other hand, the use of bounding boxes localizing pedestrians showed to have negative effects, because this approach either overloaded (visually) the scene or degraded the driver's attention on other  -- not highlighted but possibly critical -- scene elements. These outcomes indicate that, while the potential of AR for improving situational awareness is tangible, a lot of attention must be paid for the AR design to not end up cluttering and obstructing the driver's attention.

The research on augmented reality displays on windshields for improving driver awareness also extends to fully Autonomous Vehicles (AV). Such informative human-machine interfaces may help to form a mental model of the vehicle's sensory and planning system, thereby enhancing trust in AV, which is currently quite low in the general public \cite{trust1,trust2,trust3}. Lindemann \etal \cite{lindemann2018} conducted a user study on urban environments for evaluating the situational awareness of the driver in various scenarios. They found that their explanatory windshield display had positive results and improved the operator's trust. Yontem \etal \cite{Yontem2021} also designed an AR windshield interface targeting future vehicles. Their main focus was also to increase driver awareness by presenting graphical cues in a non-intrusive way based on a human-centric design and taking into account the human peripheral vision.

While the above methods provide essential feedback on the assessment of such interfaces' design, a significant limitation is that most studies were based on basic or non-interactive simulations, with the steering wheel and pedals not influencing the simulated environment and thus restricting the feeling of immersiveness of the simulations during the evaluation study. A more realistic, experimental study on the benefits of AR in driver's behavior was performed by Kim \etal \cite{kim2018} outdoors in a parking lot. It focused on pedestrian collision warning based on visual depth cues delivered in a conformal manner through a monocular display seated above the dashboard, or a volumetric display providing binocular disparity. A limitation of this study, which we address through our AR visualization component (subsection \ref{subsec:visualization} of section \ref{sec:visualization_communication}), is the limited field of view of the display used in the experiments, potentially creating a tunneling effect of the human vision.

%% file: dataset.tex
\subsection{Data simulations}
For evaluation of our methodology, we created a rich dataset using CARLA, an open-source autonomous driving simulator \cite{carla17}. CARLA is based on a server-client system, in which the server is responsible for running the simulations including the calculation of physics, weather conditions, collision detection and sensor readings. It operates on the OpenDRIVE specification \cite{opendrive} for defining junctions, traffic lights, etc, and is used by CARLA for simulating independent agents, such as other cars and pedestrians. This makes CARLA ideal for creating complex scenarios and realistic driving conditions for our tests. 

The server running the simulations is powered by Unreal Engine. Clients can connect and request changes to almost any element in the world being essential for the creation of scenarios. They also receive sensor data and manage input to the vehicle controlled by the user. CARLA supports a wide range of sensor suites with extensive configurability to its intrinsic parameters. In our work, we use a LiDAR sensor on top of the vehicle and a monocular RGB camera, placed in the front part of the car, for simulated data collection. By placing these sensors in an autonomous car and initiating its navigation in the virtual environment, we were able to create a very large dataset for evaluating our algorithms. In the future, we plan to assess the AR visualization effectiveness, with respect to reaction time and awareness increase, in a real environment with a driver manually controlling a vehicle. %in real-time.

\subsubsection*{Contributions in CARLA simulator}
Due to lack of benchmark point clouds datasets representing real road scenes with obstacles (potholes and bumps), we used the CARLA simulator to create obstacle-free environment data, in which we subsequently introduced simulated obstacles. Specifically, we designed obstacles as curved point cloud surfaces using the open-source software Blender\footnote{https://www.blender.org/} and used them to substitute parts of the road. To avoid modeling the obstacles by hand, we followed an automated procedure to generate a plethora of different obstacles based on several parameters, such as depth, ellipticity and size. An example of a frame in the CARLA simulator with a simulated pothole is presented in Fig. \ref{fig:oneframe} (texture) and in Fig. \ref{fig:saliency_mapping} (geometry).

%%% There are more that have to be described in this paragraph.

%% file: Table_Eval.tex
%\kant[1-2]

\begin{table}[htb]
  \caption{\label{table_eval} Evaluation metrics for pothole detection (in percentage \%)}
 \resizebox{\columnwidth}{!}{
 \begin{tabular}{|c|c|c|}  \hline %\begin{tabular}{|c|*{3}{>{$}c<{$}}} 
 ($\times 100\%$) & Annotated as Pothole & Annotated as Road \\ \hline
 Actual Pothole & $\text{RP} = \frac{\text{TP}}{\text{P}}$  & $\text{NR = 1 - RP}$ \\ \hline
 Actual Road & $\text{NP = 1 - RR}$  & $\text{RR} = \frac{\text{TN}}{\text{R}}$ \\ \hline
 \end{tabular}
  }
\label{tableI}
\end{table}
%\kant[3-5]

% \noindent Percentage of vertices correctly annotated as pothole, Real Pothole (\text{RP}):
% \begin{equation}
% \text{RP} = \frac{n_{TP}}{n_P} \times 100 \%
% \label{TP}
% \end{equation}
% Percentage of vertices correctly annotated as road, Real Road (\text{RR}):
% \begin{equation}
% \text{RR} = \frac{n_{TN}}{n_R} \times 100    
% \label{TR}
% \end{equation}
% Percentage of vertices wrongly annotated as pothole, Not Real Pothole ($\text{NP}$):
% \begin{equation}
% \text{NP} = \frac{n_{FP}}{n_R} \times 100 \% = 100 \% - \text{RR}
% \label{FR}
% \end{equation}
% Percentage of vertices wrongly annotated as road, not Real Road ($\text{NR}$):
% \begin{equation}
% \text{NR} = \frac{n_{FN}}{n_P} \times 100 \% = 100 \% - \text{RP}
% \label{FP}
% \end{equation}

%% file: RPCA.tex
RPCA is a powerful mathematical tool that has been used in many scientific domains in order to decompose an observed measurement $\mathbf{E}$ into a low-rank matrix $\mathbf{L}$, representing the ideal data unaffected by any kind of noise, and a sparse matrix $\mathbf{S}$, representing the noisy data. Decomposition is performed by solving the following equation:
\begin{equation}
\argmin_{\mathbf{L},\mathbf{S}} \|\mathbf{L}\|_* + \lambda \|\mathbf{S}\|_1, \ \ \
\text{s.t.} \ \mathbf{L} + \mathbf{S} = \mathbf{E},
\label{eq:decomposedrpca}
\end{equation}
where $\|\mathbf{L}\|_*$ is the nuclear norm of a matrix $\mathbf{L}$ (i.e, $\sum_i \sigma_i(\mathbf{L})$ is the sum of the singular values of $\mathbf{L}$).

A lot of works have been proposed all of these years, presenting excellent results. However, despite the effectiveness that some works \cite{candes2011robust}, \cite{Lin2009TheAL} have presented in the past, the execution times of the proposed algorithms need improvement. This convex problem can be solved using a very fast approach, as described in \cite{6738015}, according to:
\begin{equation}
%\small
\argmin_{\mathbf{L},\mathbf{S}} \frac{1}{2}\|\mathbf{L} + \mathbf{S} - \mathbf{E} \|_F + \lambda\|\mathbf{S}\|_1 \ \ \ \text{s.t.} \ \text{rank}(\mathbf{L}) = K
\label{fastpcp}
\end{equation}
\begin{equation}
%\small
\mathbf{L}^{(t+1)} = \argmin_{\mathbf{L}} \|\mathbf{L} + \mathbf{S}^{(t)} - \mathbf{E}\|_F \ \ \ \text{s.t.} \ \text{rank}(\mathbf{L}) = K
\label{update_L}
\end{equation}
\begin{equation}
%\small
\mathbf{S}^{(t+1)} = \argmin_{\mathbf{S}} \|\mathbf{L}^{(t+1)} + \mathbf{S} - \mathbf{E}\|_F + \lambda\|\mathbf{S}\|_1 
\label{update_S}
\end{equation}
In each $(t)$ iteration, the Eq. \eqref{update_L} is updated with $\text{rank} = K$. If $\frac{u_K}{\sum_{i=1}^K u_i} > \epsilon$, where $u$ denotes the singular values and $\epsilon$ is a small threshold, then the rank is increased by one (i.e., $K = K +1$) and the Eq. \eqref{update_S} is updated too. To update the Eq. \eqref{update_L}, a partial $\text{SVD}(\mathbf{E} - \mathbf{S}^{(t)})$ is estimated keeping $K$ components. To update the Eq. \eqref{update_S}, a shrinkage operator is used $\mathcal{D}(.)$, where:
\begin{equation}
%\small
\mathcal{D}( \mathbf{E} - \mathbf{L}^{(t+1)},\lambda) = \text{sign}(\mathbf{E} - \mathbf{L}^{(t+1)})\text{max}\{0,|\mathbf{E} - \mathbf{L}^{(t+1)}|-\lambda\}
\label{shrinkage_operator}
\end{equation}